\pdfoutput=1
\documentclass[letterpaper,10pt,journal,twoside]{IEEEtran}

\usepackage[pdftex]{graphicx}
\usepackage{subfigure}
\usepackage{amsmath,amssymb,amsopn,amstext,amsfonts}
\usepackage[space]{cite}
\usepackage{multicol}
\usepackage{booktabs}
\usepackage{amsthm}
\usepackage{color}
\usepackage{mathtools}

\usepackage{tabularx}
\usepackage{bm}
\usepackage{float}
\usepackage{url}
\usepackage{multirow}
\usepackage[ruled,vlined,linesnumbered]{algorithm2e}
\usepackage{pifont}
\usepackage{siunitx}
\usepackage{array}
\usepackage{xcolor}
\usepackage[hidelinks]{hyperref}
\hypersetup{
    pdftitle={DALIO: Degeneracy-Aware, Efficient LiDAR-Inertial Odometry with Adaptive Multi-Resolution Mapping},
    pdfauthor={Haoda Zhu; Fangcheng Zhu; Yuhao Fang; Jiale Han; Yunfan Ren; Siqi Liang; Longji Yin; Jie Mei; Fu Zhang}
}

\DeclareMathAlphabet\mathbfcal{OMS}{cmsy}{b}{n}

\let\oldFootnote\footnote
\newcommand\nextToken\relax

\renewcommand\footnote[1]{%
    \oldFootnote{#1}\futurelet\nextToken\isFootnote}

\newcommand\isFootnote{%
    \ifx\footnote\nextToken\textsuperscript{,}\fi}

\graphicspath{{figures/}}
\DeclareGraphicsExtensions{.pdf,.png,.jpg,.eps}

\IEEEoverridecommandlockouts
\newsavebox{\DALIOphotobox}
\newlength{\DALIOphotoheight}
\newlength{\DALIOphotowidth}
\newlength{\DALIOphotocrop}
\newcommand{\authorphoto}[2][0pt]{%
  \IfFileExists{#2}{%
    \begingroup
    \sbox{\DALIOphotobox}{\includegraphics[trim={0pt #1 0pt 0pt},clip]{#2}}%
    \setlength{\DALIOphotoheight}{\ht\DALIOphotobox}%
    \addtolength{\DALIOphotoheight}{\dp\DALIOphotobox}%
    \setlength{\DALIOphotowidth}{0.8\DALIOphotoheight}%
    \ifdim\wd\DALIOphotobox>\DALIOphotowidth
      \setlength{\DALIOphotocrop}{\wd\DALIOphotobox}%
      \addtolength{\DALIOphotocrop}{-\DALIOphotowidth}%
      \setlength{\DALIOphotocrop}{0.5\DALIOphotocrop}%
      \edef\DALIOphotooptions{trim=\the\DALIOphotocrop\space #1 \the\DALIOphotocrop\space 0pt,clip,width=1in}%
    \else
      \setlength{\DALIOphotocrop}{1.25\wd\DALIOphotobox}%
      \setlength{\DALIOphotocrop}{\dimexpr\DALIOphotoheight-\DALIOphotocrop\relax}%
      \addtolength{\DALIOphotocrop}{#1}%
      \edef\DALIOphotooptions{trim=0pt \the\DALIOphotocrop\space 0pt 0pt,clip,width=1in}%
    \fi
    \expandafter\includegraphics\expandafter[\DALIOphotooptions]{#2}%
    \endgroup
  }{%
    \PackageWarning{DALIO}{Author photo missing: #2. Not ready for final submission}%
    \begingroup
    \setlength{\fboxsep}{0pt}%
    \fbox{\parbox[c][\dimexpr1.25in-2\fboxrule\relax][c]{\dimexpr1in-2\fboxrule\relax}{%
      \centering\normalfont\footnotesize PHOTO TO BE\\PROVIDED}}%
    \endgroup
  }%
}

\title{DALIO: Degeneracy-Aware, Efficient LiDAR-Inertial Odometry with Adaptive Multi-Resolution Mapping}

\author{Haoda~Zhu$^{*}$,
        Fangcheng~Zhu$^{*}$,
        Yuhao~Fang,
        Jiale~Han,
        Yunfan~Ren,\\
        Siqi~Liang,
        Longji~Yin,
        Jie~Mei,
        and~Fu~Zhang$^\dag$
    \thanks{$^{*}$Equal contribution.}
    \thanks{$^\dag$Corresponding author: Fu Zhang (e-mail: fuzhang@hku.hk; 
phone: +852 3917 7909; 
address: HW 7--10, The University of Hong Kong, Hong Kong SAR, China).}
    \thanks{This research is supported by Smart Traffic Fund (ref: PSRI/69/2306/RA).}
    \thanks{Haoda Zhu, Fangcheng Zhu, Jiale Han, Yunfan Ren, Siqi Liang, Longji Yin, and Fu Zhang are with the Department of Mechanical Engineering, The University of Hong Kong, Hong Kong SAR, China.}
    \thanks{Yuhao Fang and Jie Mei are with the School of Intelligence Science and Engineering, Harbin Institute of Technology, Shenzhen, China.}
}

\begin{document}
\maketitle

\begin{abstract}
This paper presents DALIO, a novel LiDAR-inertial odometry (LIO) framework that achieves robust localization across diverse environments while significantly reducing computational complexity. Embodying a ``Computation on Demand" philosophy, DALIO introduces a degeneracy-aware batch selection module that dynamically adjusts the number of points used for scan registration based on real-time assessments of environmental constraints. Additionally, DALIO incorporates a degeneracy suppression method that filters out the influence of measurements along poorly constrained directions, thereby preventing noise in degenerate modes from corrupting the state update and improving overall estimation stability. Furthermore, DALIO employs MultiMap, an adaptive multi-resolution mapping strategy that preserves essential raw point data while selectively refining map resolution, effectively reducing memory usage and enabling accurate localization in both large-scale and small-scale settings. Extensive evaluations on public datasets, multiple UAV platforms, and resource-constrained embedded systems show that DALIO maintains competitive localization accuracy while reducing runtime by up to 68\% compared with FAST-LIO2 and preserving real-time performance across diverse hardware platforms.
\end{abstract}
\section{Introduction}

\IEEEPARstart{R}{ecent} advancements in LiDAR technology and the decreasing cost of sensors \cite{xiaoping2022lidar} have significantly expanded the adoption of LiDAR-inertial odometry beyond autonomous driving. Compact and affordable LiDAR sensors are now widely deployed on various mobile platforms, such as unmanned aerial vehicles (UAVs) \cite{chen2023sr} and handheld mapping equipment \cite{zheng2024livo2}, enabling precise localization in GPS-denied environments. However, these platforms typically operate under stringent hardware constraints, including limited computational resources and memory {capacity}. At the same time, they must run multiple essential modules, such as perception \cite{cai2024dmap}, planning \cite{ren2023se3}, and control \cite{liu2023ipc} in parallel with localization. This makes it critical for LIO systems to minimize computational load while maintaining high localization accuracy to ensure system responsiveness and stability.
\begin{figure}[t]
	\centering 
	\includegraphics[width=\linewidth]{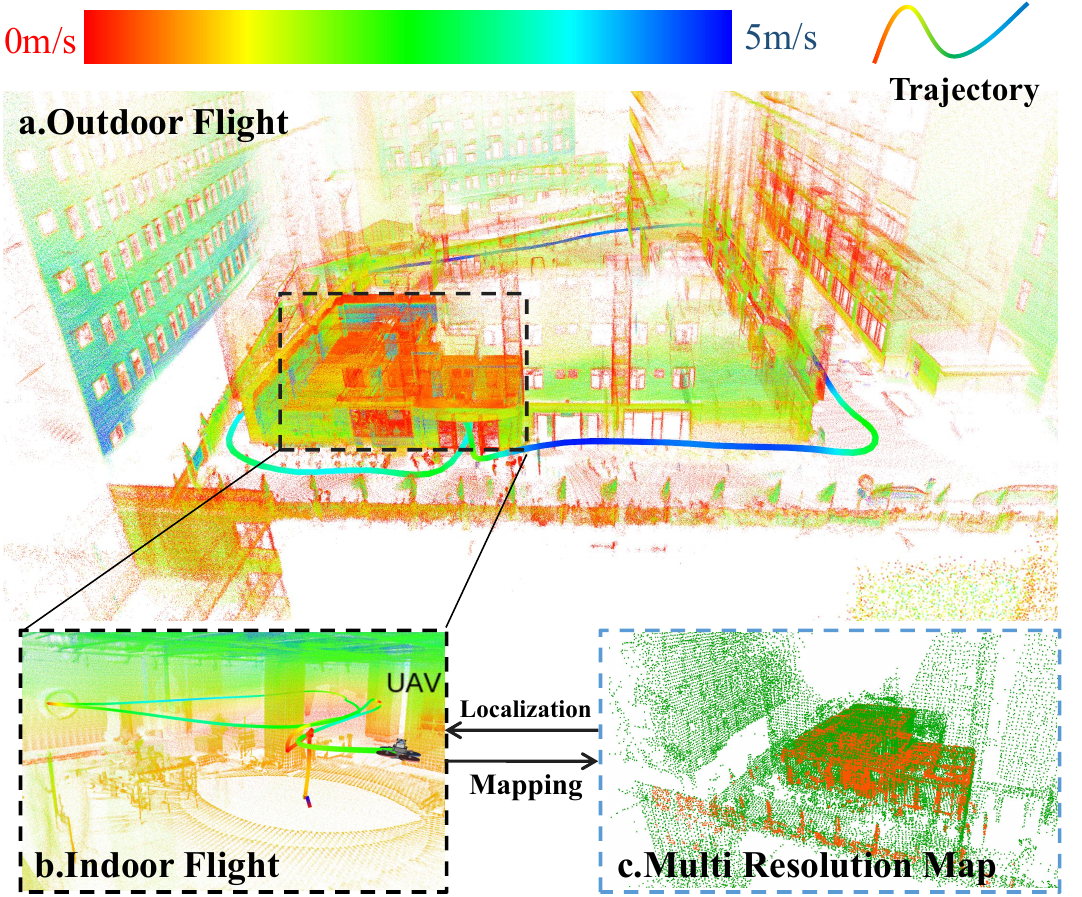}
        \caption{a. Fast autonomous navigation using DALIO on a UAV, achieving stable mapping in both indoor and outdoor environments. The trajectory is color-coded by velocity. b. Indoor flight trajectory and mapping result. c. Multi-resolution map visualization: green points indicate low resolution, red points indicate high resolution.}

	\label{fig:fly}
\end{figure}
Early methods such as LOAM \cite{zhang2014loam} and LOAM-Livox \cite{lin2020loam} employ feature extraction for real-time localization and mapping. Feature extraction and registration can nevertheless consume substantial resources on compute-limited platforms. To support real-time localization in complex and unstructured environments, recent approaches such as FAST-LIO2 \cite{xu2022fast2} eliminate explicit feature extraction to reduce computational overhead. Moreover, FAST-LIO2 uses an incremental k-d tree \cite{cai2021ikd} to enhance the efficiency of point cloud registration and map management. Additionally, FASTER-LIO \cite{gao2022faster} implements hash-based mapping structures to accelerate nearest-neighbor searches, further decreasing computational complexity. Nonetheless, several critical challenges remain unresolved:
\begin{itemize}
    \item[(1)] Current LIO methods still consume considerable computational resources, limiting onboard capacity for other essential modules. Reducing runtime and memory usage without compromising localization accuracy remains crucial for small robotic platforms.
    \item[(2)] The limited field-of-view (FOV) and mounting configurations of LiDAR sensors often cause degeneracy in point cloud registration, {leading to degraded} localization accuracy.
    \item[(3)] Existing LIO methods typically require meticulous parameter tuning due to variations between indoor and outdoor environments, limiting their general applicability and robustness.  As shown in Fig.~\ref{fig:fly}, such complex transitions remain a major challenge.

\end{itemize}

To address these challenges, we propose DALIO, a lightweight LIO framework optimized for low-cost embedded hardware and robust performance across diverse environments. Fig.~\ref{fig:fly} illustrates DALIO deployed on a UAV, achieving stable real-time localization. The key contributions of this work include:
\begin{enumerate}
\item \textit{Degeneracy-Aware Batch Selector:}
We propose a novel degeneracy-aware batch selector that segments the point cloud into multiple batches and adaptively identifies the minimal number of batches necessary for stable localization. By selectively utilizing only essential points for scan matching, this method substantially reduces computational costs while maintaining robustness under degenerate conditions.
\item \textit{Degeneracy Suppression Method:}
We introduce a degeneracy suppression method within the error-state Kalman filter (ESKF) framework that projects unreliable LiDAR sensitivity away from weakly constrained pose directions during the measurement update. This improves short-term stability and robustness when LiDAR measurements lack sufficient structure.
\item \textit{MultiMap:}
We develop an adaptive multi-resolution voxel map that adjusts resolution based on environmental geometry and prevents redundant updates in well-explored areas.  This design maintains low memory usage and consistent performance with minimal parameter tuning across diverse environments.

\item \textit{Extensive Evaluation:}
We validate DALIO using public datasets and multiple real-world platforms, including UAV flights up to \SI{8}{m/s}. Results show that DALIO maintains competitive localization accuracy with substantially lower computational cost, enabling real-time performance on low-power embedded hardware.

\end{enumerate}

\section{Related Work}

\subsection{Efficient LiDAR-Inertial Odometry}
Recent LiDAR-inertial odometry (LIO) research prioritizes computational efficiency through optimized mapping and registration. FAST-LIO2 directly registers raw points without explicit feature extraction and uses an incremental k-d tree (ikd-Tree) for map maintenance; tree rebalancing introduces additional overhead. Hash-based alternatives such as FASTER-LIO \cite{gao2022faster} and iG-LIO \cite{chen2024iglio} avoid tree rebalancing. DLIO \cite{chen2023dlio} improves lightweight direct LIO through continuous-time motion correction. LoLa-SLAM \cite{lola} reduces latency through continuous scan slicing and concurrent matching. Crucially, existing fixed-quantity point selection strategies cause excessive computation in feature-rich scenes and reduced robustness in degenerate ones. Addressing this, DALIO employs a degeneracy-aware batch selector to dynamically minimize the point batches required for robust localization, optimizing computational effort on resource-constrained platforms.

\begin{figure*}[t]
    \centering
    \includegraphics[width=1.0\textwidth]{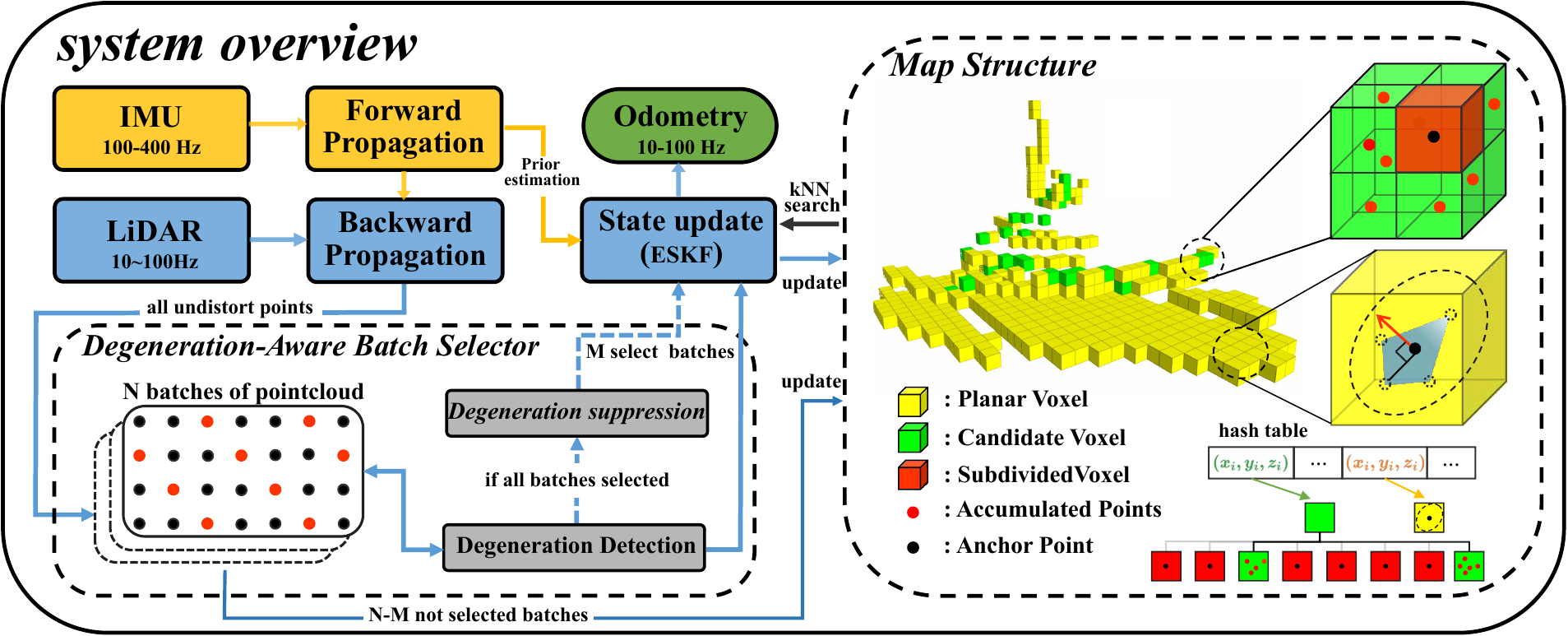}
\caption{Overview of DALIO and the MultiMap structure. The suppression branch is taken only when all batches have been processed and degeneracy persists.}
    \label{fig:system}
\end{figure*}

\subsection{Degeneracy Detection in LIO}
Degeneracy arises in geometrically featureless environments, such as corridors or open fields, where sparse constraints compromise localization. Various strategies have been proposed to mitigate this: Zhang et al. \cite{zhang2016degen} integrated degeneracy detection into the optimization framework to improve consistency; X-ICP \cite{hutter2024xicp} analyzes point cloud structure during registration to enhance robustness; Lim et al. \cite{2023adalio} adapt voxelization parameters to prevent divergence in confined spaces; and DALI-SLAM~\cite{wu2025dali} detects LiDAR feature degeneracy in its front-end odometry module and mitigates its influence through an ESKF-based update strategy. In multi-robot contexts, Zhu et al. \cite{zhu2023swarm1,zhu2025swarm2} utilized collaborative observations to compensate for single-UAV measurement degeneracy. DALIO differs from these works by coupling degeneracy analysis with computation allocation: it stops point processing once the selected batch prefix provides sufficient constraints, and applies the ESKF-compatible suppression update only when the full scan remains weakly constrained. This design reduces unnecessary computation in well-constrained frames while preserving robustness in degenerate ones.

\subsection{Multi-Resolution Map Representations}
Multi-resolution mapping balances detail preservation with memory efficiency. Hierarchical structures have been applied in reconstruction (e.g., OGN \cite{ogn2017}, HSP \cite{hsp2017}) and path planning (e.g., ROG-Map \cite{ren2024rogmap}). In the context of LIO, VoxelMap \cite{yuan2022voxelmap} combines hash tables with octrees and maintains plane parameters and their uncertainty. It supports coarse planar representations from sparse observations and limits plane updates after uncertainty converges, while retaining recent points to monitor changes.
MultiMap instead retains representative raw points and uses an initial planarity test to control subdivision. Our map-only ablation evaluates the resulting trade-off among map size, runtime, and localization accuracy.

\begin{table}[b]
	\centering
        \caption{Some Important Notations}
        \label{tab:symbols}
		\footnotesize
			\begin{tabularx}{\columnwidth}{@{}l X@{}}
			\toprule
			\textbf{Notation} & \textbf{Explanation} \\ \midrule
			$\mathbf x,\widehat {\mathbf x}, \bar{\mathbf x} ,\tilde{\mathbf x}$ & The true, predicted, updated, and error states, respectively.\\
            ${^G}\mathbf R_I$ & Orientation of the body frame $I$ in the global frame $G$.\\
            ${^G}\mathbf p_I$ & Position of the body frame $I$ in the global frame $G$.\\
            ${^G}\mathbf v_I$ & Velocity of the body frame $I$ in the global frame $G$.\\
			${^I}\mathbf R_L,{^I}\mathbf p_L,{^I}\mathbf T_L$ & The extrinsic parameters from LiDAR to IMU, where ${^I}\mathbf T_L \in SE(3)$ is the $4\times4$ transformation matrix composed of ${^I}\mathbf R_L$ and ${^I}\mathbf p_L$.\\
			${^G}\mathbf g$ & The gravity vector in the global frame. \\
			$\boldsymbol\omega_m, \mathbf a_m$ & IMU gyroscope and accelerometer measurements. \\
			$\mathbf n_g, \mathbf n_a$ & White noise of gyroscope and accelerometer measurements. \\
			$\mathbf b_g, \mathbf b_a$ & Gyroscope and accelerometer bias. \\
			$\mathbf n_{\mathbf b_g}, \mathbf n_{\mathbf b_a}$ & Gaussian noise driving the random walk of $\mathbf b_g$ and $\mathbf b_a$. \\
			\bottomrule
		    \end{tabularx}
\end{table}

\section{Framework Overview}
An overview of the DALIO system is shown in Fig.~\ref{fig:system}. For each LiDAR frame, IMU measurements are first forward-propagated to obtain a prior state estimate, which is then used to compensate for motion distortion in the point cloud. The undistorted points then enter the computation-on-demand pipeline: batch selection accumulates point batches until the localizability test is satisfied, while frames that remain weakly constrained after all batches are processed are updated through the observable-subspace suppression step. MultiMap is updated with points from the whole scan, so map maintenance is not limited to the selected registration prefix.

\section{State Estimation}
DALIO employs a tightly coupled error-state Kalman filter (ESKF) to fuse LiDAR and IMU measurements. Utilizing the notations defined in Table \ref{tab:symbols}, this section briefly outlines the state formulation based on FAST-LIO \cite{xu2021fast1}.

\subsection{State Representation and Propagation}
We adopt the manifold notation $\boxplus / \boxminus$ from \cite{hertzberg2013integrating} for the state manifold $\mathcal{M}=S O(3) \times \mathbb{R}^{15}$. The state vector $\mathbf{x} \in \mathcal{M}$ is arranged as:
\begin{equation}
    \mathbf{x} \triangleq [^{G}\mathbf R_{I}^T, \ ^{G}\mathbf p_{I}^T, \ ^{G}\mathbf v_{I}^T, \ \mathbf b_{g}^T, \ \mathbf b_{a}^T, \ ^{G}\mathbf g^T]^T
\end{equation}
Upon receiving an IMU measurement (with inputs $\boldsymbol\omega_m, \mathbf a_m$ and noise terms as listed in Table \ref{tab:symbols}), we perform forward propagation. The predicted state $\hat{\mathbf{x}}$ and covariance $\hat{\mathbf{P}}$ are updated via the standard discrete-time inertial kinematics:
\begin{equation}
\label{eq:propagation_compact}
\begin{aligned}
	\hat{\mathbf{x}}_{i+1} &= \hat{\mathbf{x}}_{i} \boxplus \left( \Delta t \cdot \mathbf{f}(\hat{\mathbf{x}}_i, \mathbf{u}_i, \mathbf{0}) \right) \\
    \hat{\mathbf{P}}_{i+1} &= \mathbf{F}_{\delta \hat{\mathbf{x}}}\, \hat{\mathbf{P}}_{i}\, \mathbf{F}_{\delta \hat{\mathbf{x}}}^T + \mathbf{F}_{\mathbf{w}} \mathbf{Q} \mathbf{F}_{\mathbf{w}}^T
\end{aligned}
\end{equation}
where $\Delta t$ is the sampling interval. The specific forms of the continuous dynamics $\mathbf{f}(\cdot)$ and the Jacobians $\mathbf{F}_{\delta \hat{\mathbf{x}}}, \mathbf{F}_{\mathbf{w}}$ follow the standard derivation in \cite{xu2021fast1}.
\subsection{State Update}

In this section, we introduce the entire process of state update, including batch selection (Section \ref{sec:batch-selector}), degeneracy analysis (Section \ref{sec:degeneracy-analysis}), and modeling of LiDAR measurements (Section \ref{sec:measurement-model}).

\begin{algorithm}[t]
\LinesNumbered
\SetAlgoLined
\caption{Degeneracy-Aware Batch Selector}
\label{alg:batch_selection}
\KwIn{Point cloud \( \mathbf{P} \), batch number \( N \), threshold \( \epsilon_d \)}
\KwOut{Selected point set \( \mathbf{S} \) for localization}

Split \( \mathbf{P} \) into \( N \) batches \( \mathbf{B}_1, \mathbf{B}_2, \ldots, \mathbf{B}_N \) of equal size\; \label{line:split_batches}
Initialize \( \mathbf{S} \gets \emptyset \), \( \mathbf{A} \gets \mathbf{0} \), \( \mathbf{A_{obs}} \gets \mathbf{0} \);

\For{\( i = 1 \) to \( N \)}{
    \( \{ \mathbf{H}_p\}_{p \in \mathbf{B}_i} = \texttt{FindNearestPlane}(\mathbf{B}_i) \)\;
    \label{line:FindNearestPlane}
    \( \mathbf{A_{obs}} = \texttt{Compute}(\{\mathbf{H}_p\}_{p \in \mathbf{B}_i}) \)\; 
    \label{line:compute_matrices}
    \( \mathbf{A} \gets \mathbf{A} + \mathbf{A_{obs}} \)\; 
    \label{line:accumulate_matrices}
    \( \lambda_{d}^{(i)} = \texttt{FindMinEigenvalue}(\texttt{SVD}(\mathbf{A})) \)\;
    \label{line:svd_accumulated}

    \If{\( \lambda_{d}^{(i)} \ge \epsilon_d \)}{ \label{line:check_accumulated}
        Set \( \mathbf{S} \gets \bigcup_{j=1}^{i} \mathbf{B}_j \)\;
        \Return \( \mathbf{S} \)\; \label{line:return_accumulated}
    }
}
\Return \( \bigcup_{i=1}^{N} \mathbf{B}_i \)\;
\end{algorithm}

\subsubsection{Degeneracy-Aware Batch Selection}\label{sec:batch-selector}
After motion compensation, we obtain an undistorted LiDAR scan. Since LIO may operate in both large-scale and small-scale environments, the number of points per frame can vary considerably. 
In large-scale scenes, even a relatively small subset of points may already provide sufficient constraints for localization due to the abundant LiDAR measurements. 
To realize ``Computation on Demand", we propose the Degeneracy-Aware Batch Selector. Instead of processing the entire scan blindly, this module avoids utilizing unnecessary points by incrementally evaluating environmental constraints.

Alg.~\ref{alg:batch_selection} performs point selection in an incremental manner.  
The undistorted point cloud is first downsampled and divided into \(N\) batches (line~\ref{line:split_batches}).  
For each batch, nearest-neighbor search and local plane fitting produce the pose-related observation terms that Algorithm~\ref{alg:batch_selection} accumulates in \ensuremath{\mathbf{A}}; see \mbox{lines~\ref{line:FindNearestPlane} to~\ref{line:accumulate_matrices}}. The minimum eigenvalue \ensuremath{\lambda_d^{(i)}} then measures the weakest constrained pose direction after the first \ensuremath{i} batches have been accumulated.
The selector returns the first batch prefix whose weakest accumulated constraint reaches \ensuremath{\epsilon_d}; see \mbox{lines~\ref{line:check_accumulated} to~\ref{line:return_accumulated}}. Therefore, point processing stops once the selected batches provide sufficient local constraints. If no prefix passes the threshold, all batches are used for the update.

The nearest-neighbor search and plane-fitting results are reused in the Kalman filter update, avoiding duplicate searches and plane fitting. Batch selection additionally requires accumulating the pose-related matrices and computing their eigenvalues.

\subsubsection{Degeneracy Analysis}\label{sec:degeneracy-analysis}
We analyze degeneracy as follows. To evaluate potential degeneracy, we extract the pose-related part of each measurement Jacobian matrix related to point ${^L}\mathbf p_j, j=1,..., m$:
\begin{equation}
    \mathbf{H}_j^{\text{pose}} \triangleq (\mathbf{H}_j)_{1:6} \in \mathbb{R}^{1\times 6},
\end{equation}
where $m$ denotes the number of planar feature points in the selected batches.
Then we construct the observability matrix by stacking these rows across all $m$ measurements:
\begin{equation}
    \mathbf{H}_{\text{obs}} = 
\begin{bmatrix}
\mathbf{H}_1^{\text{pose}} \\
\vdots \\
\mathbf{H}_m^{\text{pose}}
\end{bmatrix}
\in \mathbb{R}^{m \times 6}.
\end{equation}
We then compute the eigendecomposition of the symmetric matrix:
\begin{equation}
    \mathbf{A}_{\mathrm{obs}}=\mathbf{H}_{\mathrm{obs}}^{T}\mathbf{H}_{\mathrm{obs}} = \mathbf{V}\mathbf{\Sigma}\mathbf{V}^{T},
\label{eq:svd}
\end{equation}
where $\mathbf{V}$ contains the eigenvectors associated with the six pose parameters, $\mathbf{\Sigma} = \mathrm{diag}(\lambda_1, \lambda_2, \dots, \lambda_6)$ contains the eigenvalues, ordered as $\lambda_1 \le \lambda_2 \le \dots \le \lambda_6$. We denote the smallest eigenvalue as $\lambda_{\min} = \lambda_1$ and compare it with a predefined threshold $\epsilon_d$.
A frame is regarded as non-degenerate if $\lambda_{\min} \ge \epsilon_d$, indicating that the point cloud provides sufficient constraints for stable localization. Otherwise, if $\lambda_{\min} < \epsilon_d$, the frame is considered degenerate, in which case the degeneracy-aware batch selector activates to incorporate additional points for localization (see Section \ref{sec:batch-selector}).

\subsubsection{Measurement Model}\label{sec:measurement-model}
Assuming that after several rounds of batch selection, sufficient points in the current scan are accumulated and LiDAR degeneracy is not detected, we can proceed to state update.
Denote by $k$ the current scan index and by $\bigl\{{}^{L}\mathbf{p}_{j},\ j=1,\dots,m \bigr\}$ the planar feature points accumulated by the batch selection, each expressed in the local LiDAR frame $L$ at the scan end time. Due to LiDAR measurement noise, each measured point ${}^{L}\mathbf p_{j}$ is contaminated by noise ${}^{L}\mathbf n_{j}$, comprising range and beam-direction errors. Removing this noise yields the true point location in the LiDAR frame:

\begin{equation}
    {}^{L}\mathbf p_{j}^{\mathrm{gt}} = {}^{L}\mathbf p_{j} + {}^{L}\mathbf n_{j}.
\end{equation}

This true point, when projected into the global frame using the extrinsic ${}^{I}\mathbf T_{L}$, should lie exactly on a small plane in the map. Thus, the measurement model is

\begin{equation}
    0 = \mathbf u_{j}^{T}\left({}^{G}\mathbf T_{I}\,{}^{I}\mathbf T_{L}\left({}^{L}\mathbf p_{j} + {}^{L}\mathbf n_{j}\right) - {}^{G}\mathbf q_{j}\right) 
    \label{res_comp}
\end{equation}%
where $\mathbf u_{j}$ is the normal vector of the corresponding plane and ${}^{G}\mathbf q_{j}$ is a point on that plane. Here, a rigid transform acts on a 3-D point as $\mathbf T\mathbf p \equiv \mathbf R\mathbf p+\mathbf t$.

Approximating the implicit measurement equation (\ref{res_comp}) by its first-order Taylor expansion at $\hat{\mathbf{x}}_{k}$ yields:
\begin{align}\label{eq:linearized-residual}
0 &= h_{j}\bigl(\mathbf{x}_{k},{}^{L}\!\mathbf{n}_{j}\bigr) 
   \approx h_{j}\bigl(\hat{\mathbf{x}}_{k},\mathbf{0}\bigr)
        + \mathbf{H}_{j}\tilde{\mathbf{x}}_{k}
        + r_{j}
    \nonumber \\
  &= z_{j} + \mathbf{H}_{j} \tilde{\mathbf{x}}_{k}+ r_{j},
\end{align}
where $\mathbf{H}_{j}$ is the measurement Jacobian, which characterizes the local sensitivity of the measurement with respect to the state and determines whether the optimization problem is sufficiently constrained. Here, $z_{j}$ denotes the constant term obtained by evaluating the measurement function at the linearization point $\hat{\mathbf{x}}_{k}$ without noise, while $r_{j}$ represents the residual measurement noise after linearization, with the associated covariance matrix given by $\mathbf{R} = \mathbf{E}\{ r_{j} r_{j}^{T} \}$. The explicit form of $\mathbf{H}_{j}$'s row vector is:

\begin{equation}\label{eq:hj_row}
    \mathbf{H}_j = 
    \left[ 
        - \mathbf u_{j}^{T} {}^{G}\widehat{\mathbf{R}}_{I} \left( {}^{I}\mathbf{R}_{L}{}^{L}\mathbf{p}_{j} + {}^{I}\mathbf{p}_{L} \right)^{\wedge},\ 
        \mathbf u_{j}^{T},\ 
        \mathbf{0}_{1\times12} 
    \right]
\end{equation}
with $(\cdot)^{\wedge}$ denoting the skew-symmetric matrix operator. The rotation is evaluated at the predicted state.

Then, the error state $\tilde{\mathbf{x}}_{k}$ is estimated through the maximum a posteriori (MAP) formulation
\begin{equation}
\min_{\tilde{\mathbf{x}}_{k}}\;
\left(
\left\| \mathbf{x}_{k} \boxminus \hat{\mathbf{x}}_{k} \right\|_{\hat{\mathbf{P}}_{k}^{-1}}^{2}
+ \sum_{j=1}^{m} \left\| z_{j}^{k} + \mathbf{H}_{j}^{k}\,\tilde{\mathbf{x}}_{k} \right\|_{\mathbf{R}^{-1}}^{2}
\right),
\label{eq:eskf-map_old}
\end{equation}
where $\hat{\mathbf{x}}_{k}$ and $\hat{\mathbf{P}}_{k}$ denote the propagated state and covariance at time $k$, respectively. 
Here, $\mathbf{H}_{j}^{k}$ represents the $j$-th Jacobian used in the MAP update.

\subsection{Degeneracy Suppression in State Update}\label{sec:degeneracy-suppression}

When the system remains degenerate after using all point batches, we perform an ESKF-compatible observable-subspace measurement update by projecting the pose-related LiDAR Jacobian before it enters the MAP formulation.
$\mathbf{V}\in \mathbb{R}^{6\times 6}$ in (\ref{eq:svd}) is the orthonormal matrix of eigenvectors.
We partition $\mathbf{V}$ into two orthogonal subspaces according to their observability strength:
\begin{equation}
\mathbf{V} =
\begin{bmatrix}
\mathbf{V}_f & \mathbf{V}_u
\end{bmatrix},
\quad \mathbf{V}_f \in \mathbb{R}^{6 \times r}, \quad \mathbf{V}_u \in \mathbb{R}^{6 \times (6-r)}.
\end{equation}
Here, $\mathbf{V}_f$ contains the eigenvectors whose eigenvalues are at least the threshold $\epsilon_d$ and thus span the well-observed subspace, while $\mathbf{V}_u$ contains those below $\epsilon_d$, corresponding to noise-dominated directions.

Since $\mathbf{V}$ is orthonormal, the identity
\begin{equation}
\mathbf V\mathbf V^{T}
= \begin{bmatrix}\mathbf V_f & \mathbf V_u\end{bmatrix}
\begin{bmatrix}\mathbf V_f^{T} \\ \mathbf V_u^{T}\end{bmatrix}
= \mathbf V_f\mathbf V_f^{T} + \mathbf V_u\mathbf V_u^{T}
= \mathbf I_6
\end{equation}
confirms that $\mathbf{V}_f$ and $\mathbf{V}_u$ form two mutually orthogonal and complete subspaces.  
Therefore, any pose-related Jacobian row vector $\mathbf{H}_j^{\text{pose}} \in\mathbb{R}^{1\times 6}$ can be uniquely decomposed as
\begin{equation}
\mathbf{H}_j^{\text{pose}} = \mathbf{d}_f\mathbf{V}_f^{T} + \mathbf{d}_u\mathbf{V}_u^{T},
\end{equation}
where $\mathbf{d}_f\in\mathbb{R}^{1\times r}$ and $\mathbf{d}_u\in\mathbb{R}^{1\times(6-r)}$ are the projections onto the informative and degenerate subspaces, respectively.

Since our objective is to retain only the informative component $\mathbf{d}_f\mathbf{V}_f^{T}$ and eliminate the noise-dominated part $\mathbf{d}_u\mathbf{V}_u^{T}$, this can be achieved by operating on $\mathbf{H}_j$ using the orthogonal projection operator $\mathbf{V}_f\mathbf{V}_f^{T}$ as
\begin{equation}
\begin{aligned}
\bar{\mathbf{H}}_j^{\text{pose}}
&= \mathbf{H}_j^{\text{pose}}\mathbf{V}_f\mathbf{V}_f^{T} 
= (\mathbf{d}_f\mathbf{V}_f^{T} + \mathbf{d}_u\mathbf{V}_u^{T})\mathbf{V}_f\mathbf{V}_f^{T} \\
&= \mathbf{d}_f\mathbf{V}_f^{T}\mathbf{V}_f\mathbf{V}_f^{T} + \mathbf{d}_u\mathbf{V}_u^{T}\mathbf{V}_f\mathbf{V}_f^{T} 
= \mathbf{d}_f\mathbf{V}_f^{T},
\end{aligned}
\end{equation}
and the complementary component is given by
\begin{equation}
\tilde{\mathbf{H}}_j^{\text{pose}} 
= \mathbf{H}_j^{\text{pose}} - \bar{\mathbf{H}}_j^{\text{pose}} 
= \mathbf{d}_u\mathbf{V}_u^{T}.
\end{equation}

Applying this projection to all rows, we obtain the filtered observability matrix:
\begin{equation}
\bar{\mathbf{H}}_{\mathrm{obs}}
= \begin{bmatrix}
  \bar{\mathbf{H}}_1^{\text{pose}}\\
  \vdots\\
  \bar{\mathbf{H}}_m^{\text{pose}}
\end{bmatrix}
= \mathbf{H}_{\mathrm{obs}}\;\mathbf{V}_f \mathbf{V}_f^{T}.
\end{equation}
Let $\bar{\mathbf{H}}_j^{\text{pose}}$ replace the first six pose-related columns of $\mathbf{H}_j$, while the remaining non-pose columns remain unchanged. The complementary matrix corresponding to $\mathbf{H}_j$ is then given by
\begin{equation}
\tilde{\mathbf{H}}_j = \mathbf{H}_j - \bar{\mathbf{H}}_j = [\,\tilde{\mathbf{H}}_j^{\text{pose}},\, \mathbf{0}_{1 \times 12}\,].
\end{equation}
To apply the filtered matrix $\bar{\mathbf{H}}_{\mathrm{obs}}$ in the MAP formulation~\eqref{eq:eskf-map_old}, the linearized residual equation~\eqref{eq:linearized-residual} needs to be reformulated as
\begin{equation}
\begin{aligned}
0 &= z_{j} + \mathbf{H}_{j}\tilde{\mathbf{x}}_{k} + r_{j} \\
&= z_{j} + (\bar{\mathbf{H}}_{j} + \tilde{\mathbf{H}}_j) \tilde{\mathbf{x}}_{k} + r_{j} \\
&= z_{j} + \bar{\mathbf{H}}_j \tilde{\mathbf{x}}_{k}+ \tilde{\mathbf{H}}_j \tilde{\mathbf{x}}_{k} + r_{j} \\
&= z_{j} + \bar{\mathbf{H}}_{j} \tilde{\mathbf{x}}_{k} + \tilde{r}_j,
\end{aligned}
\end{equation}
where $\tilde{r}_j$ denotes the newly modeled observation noise,
\begin{equation}
\begin{aligned}
    \tilde{r}_j &= \tilde{\mathbf{H}}_j \tilde{\mathbf{x}}_{k} + r_{j}.
\end{aligned}
\end{equation}
Since $\tilde{\mathbf{x}}_{k}$ and $\mathbf{r}_j$ are generally assumed to be uncorrelated, the covariance matrix of $\tilde{r}_j$ can be computed as
\begin{equation}
\begin{aligned}
    \tilde{\mathbf{R}}_j 
    &= \mathbf{E}\{\tilde{r}_j \tilde{r}_j^T\} \\
    &= \mathbf{E}\{(\tilde{\mathbf{H}}_j \tilde{\mathbf{x}}_{k}) (\tilde{\mathbf{H}}_j \tilde{\mathbf{x}}_{k})^T \} + \mathbf{E}\{ r_{j} r_{j}^{T} \} \\
    &= \tilde{\mathbf{H}}_j \mathbf{E}\{ \tilde{\mathbf{x}}_{k} \tilde{\mathbf{x}}_{k}^T \} \tilde{\mathbf{H}}_j^T + \mathbf{R} \\
    &= \tilde{\mathbf{H}}_j \hat{\mathbf{P}}_k \tilde{\mathbf{H}}_j^T + \mathbf{R}.
\end{aligned}
\end{equation}
Thus, the reformulated residual remains closed in the same linearized MAP problem, with the removed sensitivity represented by \ensuremath{\tilde{\mathbf{R}}_j}.
Then, the error state $\tilde{\mathbf{x}}_{k}$ can be estimated using the new MAP formulation
\begin{equation}
\min_{\tilde{\mathbf{x}}_{k}}\;
\left(
\left\| \mathbf{x}_{k} \boxminus \hat{\mathbf{x}}_{k} \right\|_{\hat{\mathbf{P}}_{k}^{-1}}^{2}
+ \sum_{j=1}^{m} \left\| z_{j}^{k} + \bar{\mathbf{H}}_{j}^{k}\tilde{\mathbf{x}}_{k} \right\|_{\tilde{\mathbf{R}}_j^{-1}}^{2}
\right).
\label{eq:eskf-map}
\end{equation}

The underlying nonlinear estimation problem is solved by repeated linearization. Each iteration forms a local quadratic problem in the tangent space, following the iterated Kalman-filter/Gauss--Newton formulation \cite{ikf_gs}.
To respect the manifold structure $\mathcal{M}$ of the state, each iteration parameterizes the update in the tangent space (i.e., the error state) around the current estimate. 
This inner loop continues until convergence. 
The converged state, denoted as $\bar{\mathbf{x}}_{k}$, and the corresponding covariance $\bar{\mathbf{P}}_{k}$ obtained from the Hessian of (\ref{eq:eskf-map}), are subsequently used for IMU propagation and for updating the global LiDAR map.

Crucially, the projected Jacobian \ensuremath{\bar{\mathbf{H}}_{\mathrm{obs}}=\mathbf{H}_{\mathrm{obs}}\mathbf{V}_f\mathbf{V}_f^{T}} restricts measurement sensitivity to the informative subspace spanned by \ensuremath{\mathbf{V}_f}. The update therefore injects information only along well-observed directions, while the covariance remains unchanged in the degenerate subspace spanned by \ensuremath{\mathbf{V}_u}. This keeps the filter conservative during weak-constraint intervals and prevents LiDAR noise from producing overconfident state corrections.

\subsection{Map Design}\label{section:MultiMap}

The degeneracy-aware batch selector adapts the number of points processed for registration to the available geometric constraints. To maintain stable performance and consistent computational efficiency across various scenarios, we propose a multi-resolution map structure called MultiMap.

\subsubsection{Data Structure of MultiMap}

The proposed map utilizes a hierarchical, hash-based storage approach. Specifically, point clouds are initially stored in sparse voxels, whose indices are hashed into an unordered map. Each voxel contains an octree structure with a predefined maximum depth, facilitating efficient spatial subdivision.

To effectively control memory usage, we introduce a Least Recently Used (LRU) caching algorithm. This algorithm maintains a fixed-length doubly-linked list to record the usage order of voxel nodes. Whenever a voxel node is accessed or updated, it is moved to the front of the list, indicating recent usage. If the number of voxel nodes exceeds the predefined capacity, the node at the tail of the list---the least recently used node---is removed from both the linked list and the hash table, thereby releasing its memory.

\subsubsection{Initialization and Update of MultiMap}

When a new frame of point cloud data arrives, each point is first mapped to the coarsest-resolution voxel. 
If the voxel does not exist, it is created and initialized. 
In either case, the point is inserted into the \textit{Accumulated Points} set $\mathcal{P}_{acc}$.

Each voxel maintains an internal octree whose nodes are classified into three states: \textit{Candidate Voxel}, \textit{Planar Voxel}, or \textit{Subdivided Voxel}, as shown in Fig.~\ref{fig:system}. 
At each octree level, a voxel is subdivided into eight child voxels. 
When at least four of these sub-voxels have accumulated points, a PCA is performed over all points in $\mathcal{P}_{acc}$ to assess planarity:
\[
\mathbf{C} = \frac{1}{|\mathcal{P}_{acc}|} \sum_{p \in \mathcal{P}_{acc}} (p - \bar{p})(p - \bar{p})^\top,
\]
where $\bar{p}$ denotes the centroid of $\mathcal{P}_{acc}$ and $\mathbf{C}$ is its sample covariance matrix. 
Let $\lambda_1 \geq \lambda_2 \geq \lambda_3$ be the eigenvalues of $\mathbf{C}$. 
If the smallest eigenvalue ratio $\lambda_3 / (\lambda_1 + \lambda_2 + \lambda_3)$ is below a threshold $\tau_{\mathrm{pln}}$, the voxel is marked as a \textit{Planar Voxel}, and only the point closest to the voxel center---the \textit{Anchor Point} $p_{anc}$---is retained. Here, PCA serves as a map-resolution test: a low ratio means that the points in the voxel can be represented at the current coarse resolution, so MultiMap keeps only the anchor point and stops further subdivision.
Otherwise, the voxel is subdivided, and each child becomes a new \textit{Candidate Voxel}. This preserves local details in rough, small-scale, or non-planar regions instead of forcing them into the same coarse representation.

\begin{figure}[t]
	\centering 
	\includegraphics[width=\linewidth]{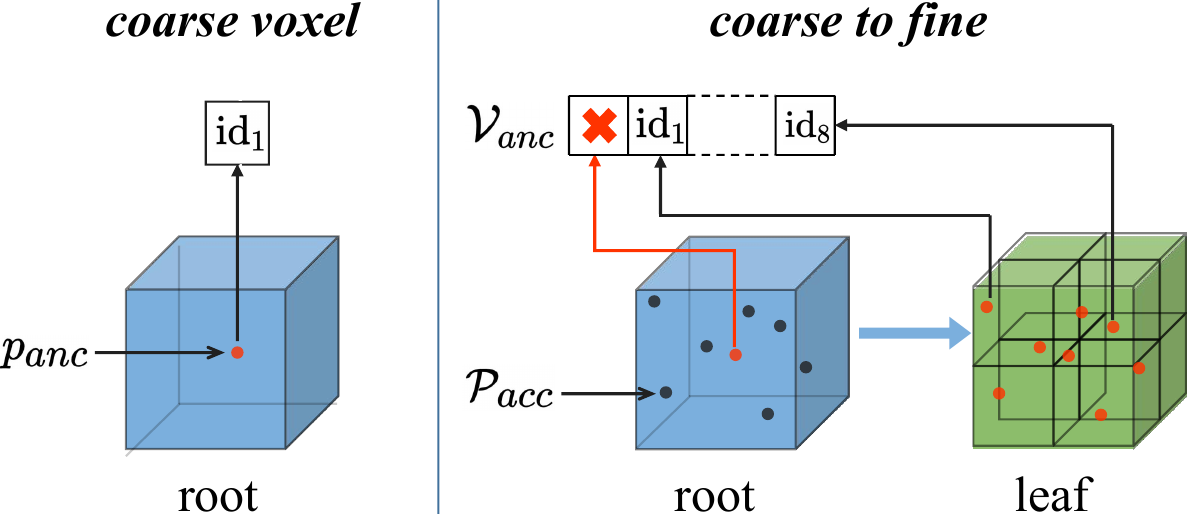}
	\caption{Anchor Point vector $\mathcal{V}_{anc}$ stores direct references to anchor points within the voxel's octree, enabling fast kNN queries with minimal traversal.}
	\label{fig:pointer}
\end{figure}
\begin{figure}[h]
	\centering 
	\includegraphics[width=\linewidth]{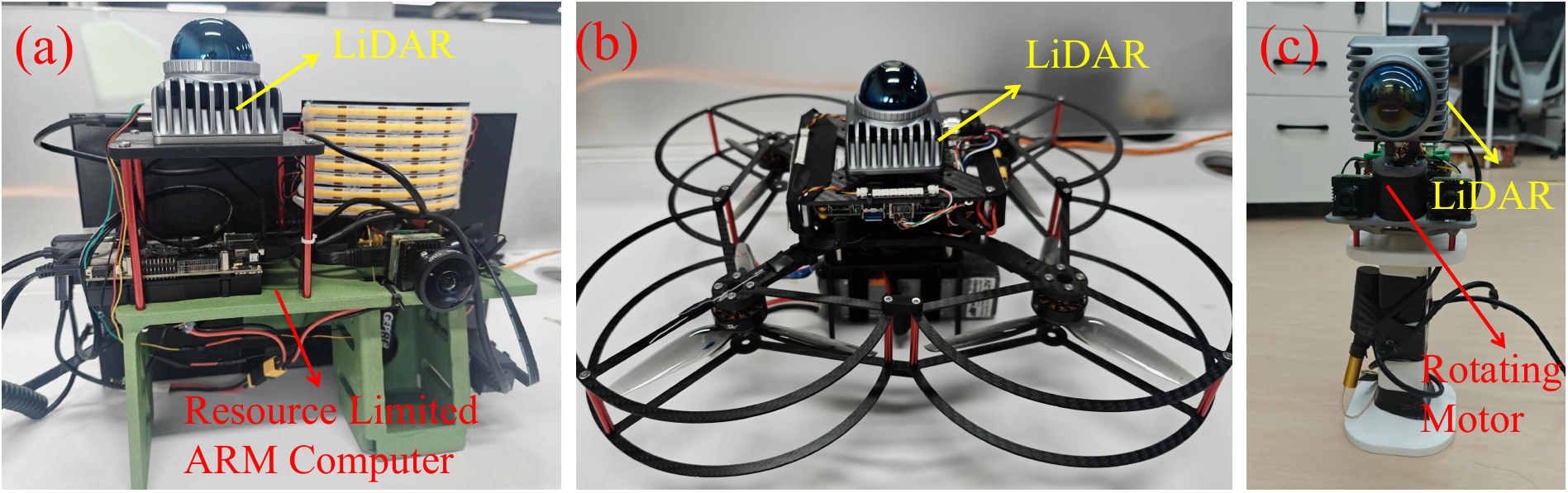}
	\caption{Three platforms used in the experiments: a handheld device with a low-cost RK3588 ARM board (approximately \$100), a UAV, and a rotating LiDAR device.}
	\label{fig:hard}
\end{figure}

This recursive process continues until reaching the maximum octree depth. At the final level, PCA is no longer applied. Once enough points are accumulated, the voxel is marked as a \textit{Subdivided Voxel}, and only $p_{anc}$ is kept. The anchor points from both \textit{Planar Voxel}s and \textit{Subdivided Voxel}s are used for subsequent kNN search.

This design is also robust to dynamic obstacles. Transient objects generally leave only sparse returns in coarse Candidate Voxels, which rarely provide enough consistent neighbors for local plane fitting and are therefore unlikely to affect the ESKF update; these inactive voxels are eventually removed by the LRU policy when the map reaches its capacity.

\subsubsection{kNN Search in MultiMap}\label{sec:knnsearch}
For k-nearest neighbor (kNN) search, each query point \(p\) is mapped to a voxel in a hash-indexed grid. We then gather points from that voxel and its adjacent voxels, compute their distances to \(p\), and return the \(K\) closest as the kNN result.

To reduce computation, each voxel maintains an auxiliary pointer vector \(\mathcal{V}_{anc}\) at the root of the octree, shown in Fig.~\ref{fig:pointer}. This vector stores direct references to all Anchor Points within the voxel's octree nodes. By limiting neighbor search to these representative points, query-time complexity is reduced to approximately \(O(1)\), shifting most computation to the map update stage.

\begin{table*}[!t]
\setlength{\tabcolsep}{2.4pt}
\renewcommand{\arraystretch}{1}
\caption{Runtime (\si{ms}), RMSE (\si{m}), and DALIO Internal Statistics for Ground-Truth Sequences}
\label{tab:time_eval_rmse}
\centering
\footnotesize
\begin{tabular*}{\textwidth}{@{\extracolsep{\fill}}l c c c c c c c c c c@{}}
\toprule[1.5pt]
\multirow{2}{*}{Seq.} & \multicolumn{3}{c}{DALIO} & DALIO$^\dagger$ & DALIO$^*$ & FAST-LIO2 & FASTER-LIO & VoxelMap & DLIO & DALI-SLAM \\
\cmidrule(lr){2-11}
 & RMSE / Time & \shortstack{Batch\\count} & \shortstack{Split\\Voxel Ratio} & RMSE / Time & RMSE / Time & RMSE / Time & RMSE / Time & RMSE / Time & RMSE / Time & RMSE / Time \\
\midrule
nclt1 & 1.34 / \textbf{10.65} & 2.12 & 0.22 & 1.35 / 10.98 & 1.31 / 24.06 & \textbf{1.21} / 31.34 & 1.24 / 20.33 & 1.28 / 22.38 & 1.26 / 20.76 & 1.48 / 32.20 \\
nclt2 & 1.56 / 9.93 & 2.46 & 0.24 & 1.55 / \textbf{9.69} & 1.61 / 26.62 & 2.03 / 31.80 & 1.59 / 19.57 & \textbf{1.44} / 23.51 & 1.62 / 19.88 & 2.14 / 32.65 \\
nclt3 & 0.98 / \textbf{11.10} & 2.91 & 0.23 & 1.00 / 11.47 & 0.92 / 25.04 & \textbf{0.88} / 29.48 & 1.03 / 17.09 & 0.99 / 23.14 & 1.05 / 17.34 & 0.94 / 30.41 \\
ncd1 & 0.33 / 16.15 & 1.95 & 0.26 & 0.34 / \textbf{15.62} & 0.32 / 30.80 & 0.35 / 36.11 & 0.32 / 28.39 & \textbf{0.31} / 26.31 & 0.33 / 28.12 & 0.37 / 36.86 \\
ncd2 & 0.32 / \textbf{17.84} & 2.17 & 0.18 & 0.31 / 18.41 & 0.35 / 30.45 & 0.35 / 37.75 & 0.38 / 31.51 & 0.33 / 30.34 & \textbf{0.30} / 31.04 & 0.36 / 38.22 \\
ncd3 & 0.12 / 15.83 & 2.09 & 0.17 & 0.12 / \textbf{15.36} & 0.12 / 27.16 & \textbf{0.11} / 32.79 & 0.14 / 26.75 & 0.13 / 29.06 & 0.13 / 26.82 & 0.12 / 33.40 \\
m3o1 & \textbf{0.20} / 8.64 & 2.84 & 0.29 & 0.21 / \textbf{8.29} & 0.21 / 14.72 & 0.23 / 16.52 & 0.25 / 14.37 & 0.24 / 18.60 & 0.26 / 10.96 & 0.22 / 15.64 \\
m3o2 & \textbf{0.19} / \textbf{11.52} & 3.36 & 0.3 & 0.20 / 11.90 & \textbf{0.19} / 16.72 & 0.30 / 29.08 & 0.20 / 16.04 & 0.94 / 29.80 & 0.21 / 16.76 & 0.28 / 28.44 \\
\bottomrule[1.5pt]
\end{tabular*}
\par\vspace{1pt}
\footnotesize
\begin{minipage}[t]{\textwidth}
\rule{0pt}{\baselineskip}Bold indicates the lowest value of each error or runtime metric in a row (including ties). DALIO$^{\dagger}$ disables degeneracy suppression; DALIO$^{*}$ uses all points for registration.
\end{minipage}
\end{table*}

\begin{table*}[!t]
\setlength{\tabcolsep}{2.4pt}
\renewcommand{\arraystretch}{1}
\caption{Runtime (\si{ms}), E2E (\si{m}), and DALIO Internal Statistics for Loop-Return Sequences}
\label{tab:time_eval_e2e}
\centering
\footnotesize
\begin{tabular*}{\textwidth}{@{\extracolsep{\fill}}l c c c c c c c c c c@{}}
\toprule[1.5pt]
\multirow{2}{*}{Seq.} & \multicolumn{3}{c}{DALIO} &DALIO$^\dagger$ & DALIO$^*$ & FAST-LIO2 & FASTER-LIO & VoxelMap & DLIO & DALI-SLAM \\
\cmidrule(lr){2-11}
 & E2E / Time & \shortstack{Batch\\count} & \shortstack{Split\\Voxel Ratio} & E2E / Time & E2E / Time & E2E / Time & E2E / Time & E2E / Time & E2E / Time & E2E / Time \\
\midrule
hku1 & 0.04 / \textbf{7.05} & 3.75 & 0.28 & 0.04 / 7.31 & 0.04 / 11.84 & \textbf{0.03} / 12.26 & 0.11 / 10.45 & 0.04 / 16.57 & 0.12 / 10.78 & 0.05 / 12.70 \\
hku2 & 0.04 / 8.58 & 3.01 & 0.36 & 0.06 / \textbf{8.33} & \textbf{0.03} / 14.39 & 0.04 / 11.83 & 0.04 / 12.39 & 0.04 / 15.44 & 0.05 / 12.71 & 0.09 / 12.31 \\
hku3 & \textbf{0.03} / \textbf{7.39} & 5.11 & 0.39 & \textbf{0.03} / 7.67 & 0.04 / 11.28 & 1.96 / 13.39 & 2.05 / 11.64 & 0.10 / 14.91 & 2.12 / 11.97 & 1.88 / 13.72 \\
ust1 & \textbf{0.65} / \textbf{9.12} & 2.91 & 0.33 & 0.91 / 9.55 & 0.93 / 15.52 & 3.38 / 16.78 & 3.87 / 14.73 & 4.91 / 24.21 & 3.96 / 15.10 & 3.52 / 17.20 \\
ust2 & 1.32 / 8.98 & 2.55 & 0.19 & 1.17 / \textbf{8.73} & \textbf{0.97} / 14.79 & 2.93 / 16.67 & 2.77 / 15.03 & 3.44 / 24.30 & 2.84 / 15.48 & 3.10 / 17.05 \\
ust3 & 0.04 / \textbf{8.08} & 3.02 & 0.27 & 0.05 / 8.41 & \textbf{0.02} / 13.59 & 0.03 / 15.25 & 0.04 / 12.02 & 0.04 / 24.60 & 0.05 / 12.36 & 0.04 / 15.82 \\
m3i1 & \textbf{0.04} / \textbf{6.72} & 8.47 & 0.3 & 0.56 / 6.93 & 0.54 / 8.60 & 1.03 / 11.01 & 0.98 / 8.89 & 3.75 / 13.31 & 1.04 / 10.11 & 1.08 / 12.44 \\
m3i2 & \textbf{0.45} / 5.49 & 7.2 & 0.32 & 1.86 / \textbf{5.36} & \textbf{0.45} / 8.55 & 3.43 / 10.84 & 5.53 / 9.29 & -- & 0.68 / 8.48 & 1.36 / 10.10 \\
m3i3 & 0.05 / \textbf{7.97} & 3.16 & 0.24 & 0.06 / 8.24 & 0.04 / 16.00 & \textbf{0.03} / 21.27 & 0.04 / 13.97 & 0.04 / 30.01 & 0.05 / 14.35 & 0.04 / 21.80 \\
dege1 & \textbf{1.01} / 8.24 & 6.45 & 0.38 & 1.58 / \textbf{8.00} & 7.35 / 10.44 & 3.13 / 18.14 & 2.28 / 9.36 & 12.85 / 24.21 & 7.41 / 10.62 & 7.02 / 19.12 \\
dege2 & \textbf{1.32} / \textbf{8.62} & 7.15 & 0.42 & 8.12 / 8.96 & 5.86 / 13.36 & 6.45 / 15.58 & 15.82 / 12.54 & -- & 5.90 / 15.82 & 6.58 / 16.38 \\
self1 & \textbf{0.06} / 9.96 & 4.38 & 0.26 & 0.09 / \textbf{9.72} & 0.08 / 18.36 & 0.11 / 19.22 & 0.07 / 15.98 & 0.09 / 21.74 & 0.08 / 16.16 & 0.12 / 19.72 \\
self2 & \textbf{0.04} / \textbf{11.76} & 6.07 & 0.49 & 2.62 / 12.23 & 0.16 / 23.94 & 9.51 / 25.14 & -- & -- & -- & 9.51 / 27.71 \\
self3 & 0.02 / 10.04 & 4.41 & 0.34 & 0.60 / \textbf{9.70} & 0.05 / 16.20 & 1.09 / 20.80 & 0.29 / 13.92 & \textbf{0.01} / 20.76 & 0.11 / 14.08 & 1.03 / 17.32 \\
self4 & 0.03 / \textbf{9.84} & 3.83 & 0.59 & 0.17 / 10.18 & \textbf{0.02} / 13.88 & 0.27 / 17.06 & 0.44 / 13.66 & 4.35 / 19.94 & 0.06 / 13.84 & 0.56 / 18.62 \\
\bottomrule[1.5pt]
\end{tabular*}
\par\vspace{1pt}
\footnotesize
\begin{minipage}[t]{\textwidth}
\rule{0pt}{\baselineskip}Variant definitions and boldface follow Table~\ref{tab:time_eval_rmse}.
\end{minipage}
\end{table*}

\section{Experiments}\label{experiments}

In this section, we extensively evaluate the proposed DALIO framework on public datasets, as well as on handheld devices with limited computational resources, multiple UAV platforms, and a rotating LiDAR device, as illustrated in Fig.~\ref{fig:hard}. The results demonstrate that DALIO is robust and adaptable across diverse scenarios and hardware platforms, while maintaining low computational overhead.

\subsection{Evaluation on Public Datasets}

We evaluate DALIO on public datasets and self-collected robotic sequences with complementary scene properties.
The nclt~\cite{nclt-2015a} and ncd~\cite{ramezani2020newer} sequences provide large outdoor ground-truth trajectories, while hku and ust sequences from R3LIVE~\cite{lin2021r3live} cover indoor--outdoor transitions and mixed-scale structures.
In Tables~\ref{tab:time_eval_rmse} and~\ref{tab:time_eval_e2e}, m3o1--m3o2 and m3i1--m3i3 denote M3DGR~\cite{zhang2025towards} sequences, dege1--dege2 denote strongly degenerate R3LIVE sequences, and self1--self4 correspond, in the same order as the following experimental subsections, to the four self-collected experiments: RK3588 handheld mapping, rotating-LiDAR mapping, mixed indoor--outdoor UAV trajectory tracking, and high-speed forest flight.

We compare our method against FAST-LIO2, FASTER-LIO, VoxelMap, DLIO, and DALI-SLAM~\cite{wu2025dali}. For DALI-SLAM, we use only its front-end odometry module, which is the part reported to perform degeneracy detection and suppression. We also include DALIO\ensuremath{{}^{*}}, which disables the degeneracy-aware batch selector and uses all points for registration, and DALIO\ensuremath{{}^{\dagger}}, which disables only the degeneracy suppression update.

\begin{figure}[t]
	\centering 
	\includegraphics[width=\linewidth]{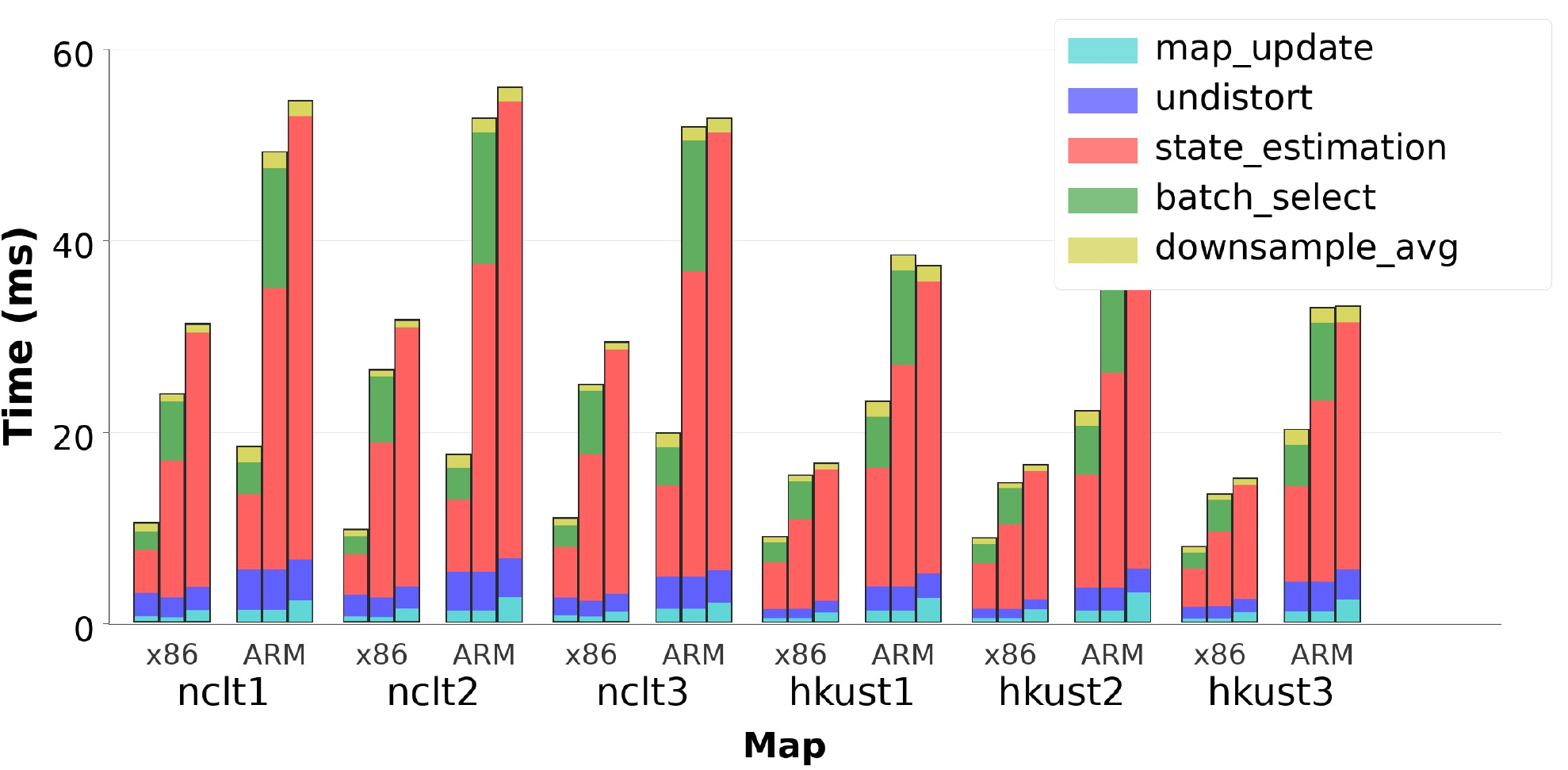}
		\caption{Module-level runtime comparison on x86 and RK3588 ARM platforms. For each dataset and platform, the three stacked bars represent DALIO, DALIO\ensuremath{{}^{*}}, and FAST-LIO2 from left to right. The x86 stacks are proportionally rescaled to the totals in Tables~\ref{tab:time_eval_rmse} and~\ref{tab:time_eval_e2e} while retaining the original component ratios.}
	\label{fig:time}
\end{figure}
In our implementation, we use a multi-resolution map with resolutions of {\SI{0.5}{m}}, {\SI{0.25}{m}}, and {\SI{0.125}{m}}. These values were chosen so that the coarsest resolution ({\SI{0.5}{m}}) matches the map resolution used in other methods. To ensure a fair comparison, parameters with similar meanings across different methods were set to the same values, while other settings followed each method's recommended configuration. All main experiments use the same default degeneracy threshold \ensuremath{\epsilon_d=20}, without per-sequence tuning. The MultiMap planarity threshold \ensuremath{\tau_{\mathrm{pln}}} is applied to the normalized smallest covariance eigenvalue in the PCA test and is set to \ensuremath{0.1} for all experiments. Split Voxel Ratio denotes the fraction of voxels that have been subdivided in MultiMap and serves as a scene-structure indicator: lower values generally correspond to large, regular outdoor scenes, whereas higher values indicate mixed-scale or locally complex scenes that retain more fine-resolution anchors. The table results are measured on an x86 workstation with an Intel i7-10875H CPU, while Fig.~\ref{fig:time} additionally compares module-level runtime on the RK3588 ARM board shown in Fig.~\ref{fig:hard}a.

\subsubsection{Efficiency Evaluation}

As shown in Table~\ref{tab:time_eval_rmse} and Table~\ref{tab:time_eval_e2e}, DALIO and DALIO$^{\dagger}$ consistently achieve the lowest computational cost among all compared methods. The consistently higher computation time of DALIO$^*$ highlights the effectiveness of the proposed batch selector in reducing processing cost. Because the same default threshold is used across all main experiments, the lower runtime comes from frame-wise adaptive batch allocation rather than sequence-specific parameter tuning.

\begin{figure*}[t]
		\centering
		\includegraphics[width=\linewidth]{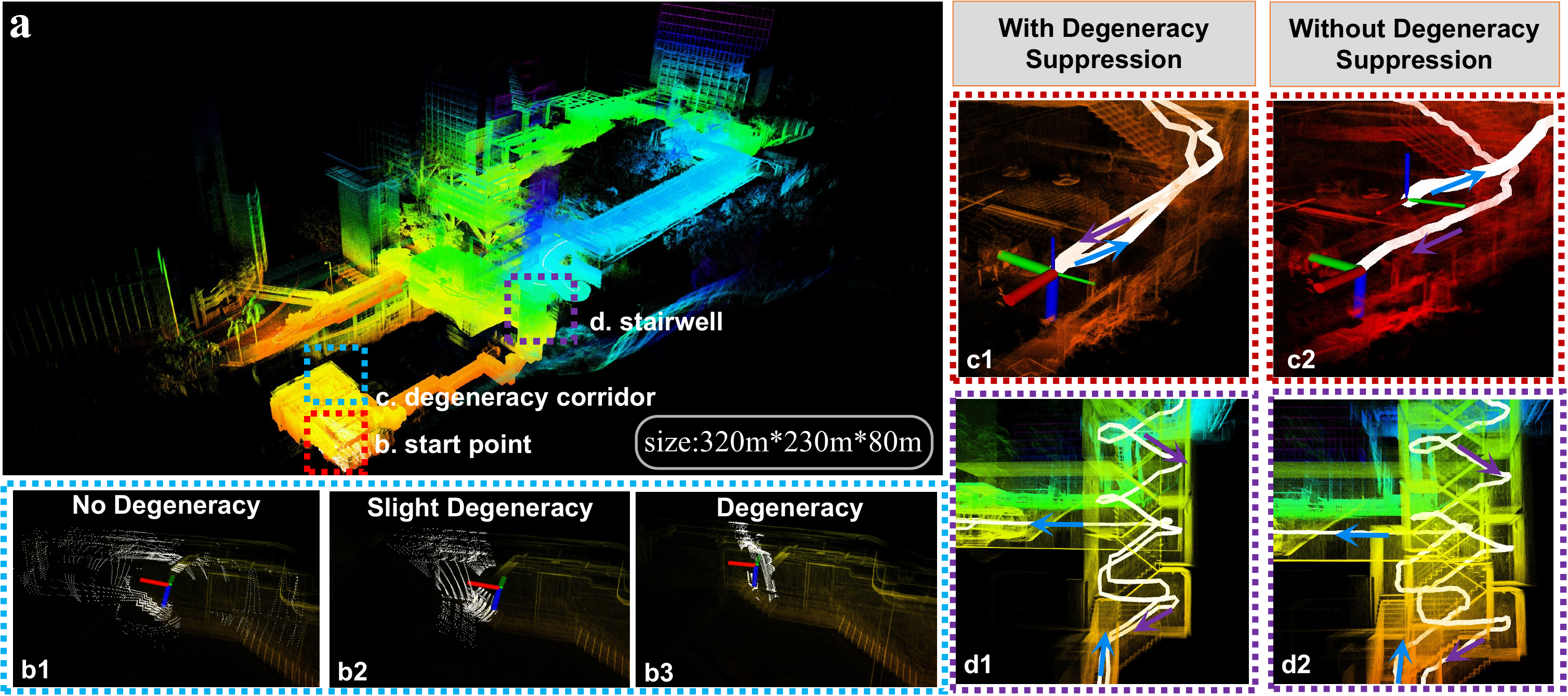}
		\caption{Fig. a shows the mapping results of the rotating LiDAR, while Figs. b1, b2, and b3 display its point clouds in narrow channels---point cloud degeneracy occurs as the LiDAR rotates. Figs. c and d compare mapping results with and without degeneracy suppression: without suppression, the point clouds show layering at the stairs, and the estimated trajectory does not return to its starting point.} 
		\label{fig:hkuBigScene} 
\end{figure*} 

\begin{figure}[t]
		\centering 
		\includegraphics[width=\linewidth]{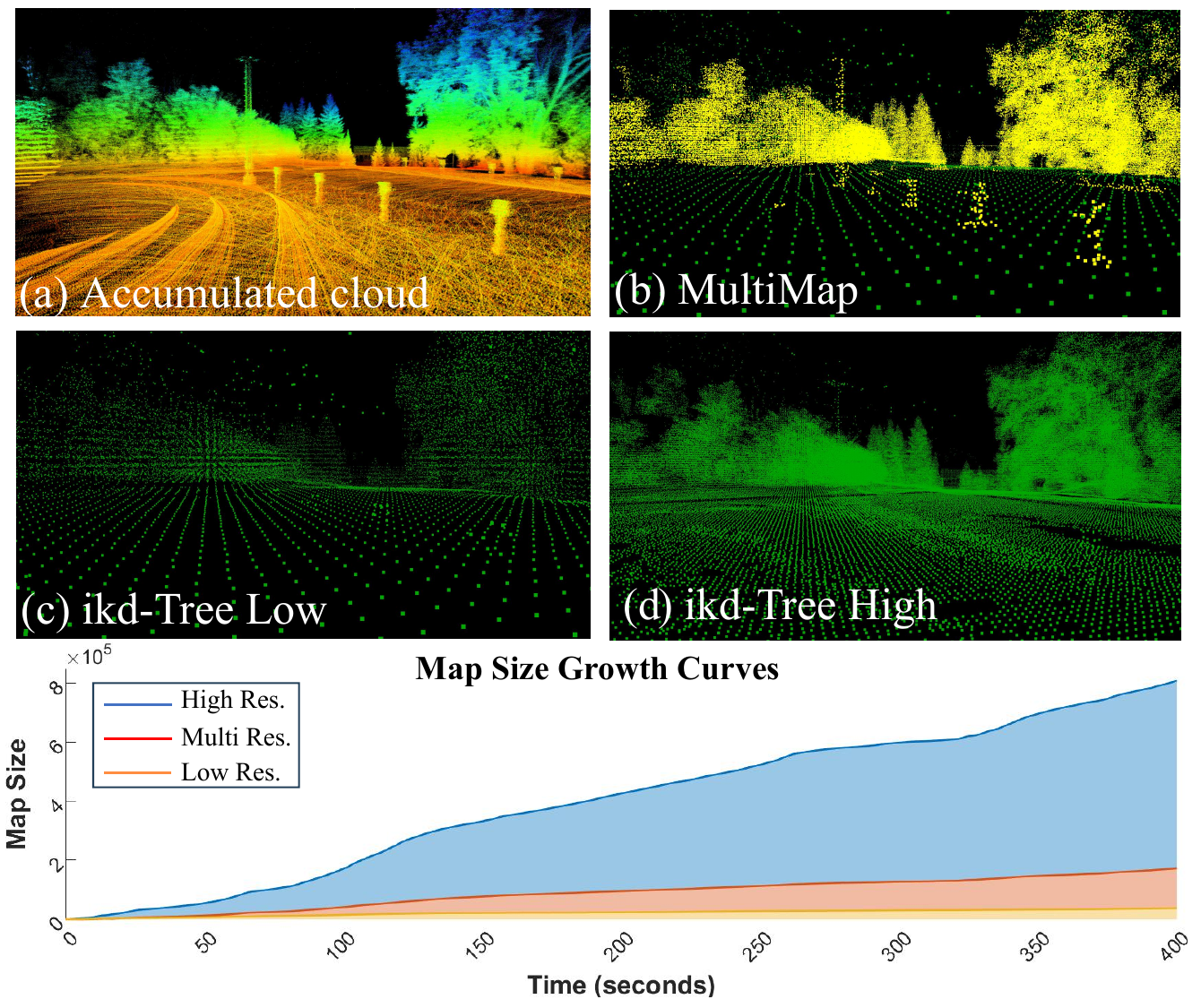}
	        \caption{Map-only ablation: (a) accumulated point cloud, (b) MultiMap, (c) ikd-Tree Low, and (d) ikd-Tree High. The lower panel shows map-size growth.}
		\label{fig:map_compare}
\end{figure}

Fig.~\ref{fig:time} further decomposes the runtime on both x86 and RK3588 ARM platforms using the same representative sequences. The lower-level modules, such as point-cloud undistortion and downsampling, remain similar across methods, whereas the main differences appear in the point-processing part of the update, including batch selection, nearest-neighbor search, and plane fitting. DALIO reduces this part by stopping once the selected batches provide sufficient local constraints. The gap becomes more pronounced on the RK3588 ARM platform, where limited computational capacity magnifies the cost of processing redundant points. This shows that the proposed computation reduction is particularly important for low-cost embedded deployment, where localization must share onboard resources with mapping, planning, and control modules.
\subsubsection{Accuracy and Robustness Evaluation}

On the ground-truth sequences in Table~\ref{tab:time_eval_rmse}, DALIO achieves RMSE close to the strongest baselines while consistently lowering runtime. The low Split Voxel Ratio on most NCLT/NCD rows indicates structured, well-constrained environments; therefore, removing the suppression update in DALIO\ensuremath{{}^{\dagger}} changes the RMSE only slightly. 

\begin{table}[!h]
\caption{Map-Only Ablation (Average E2E (\si{m}) / Average Runtime (\si{ms}))}
\label{tab:hku_hkust_e2e_time}
\centering
\footnotesize
\setlength{\tabcolsep}{3.5pt}
\begin{tabular*}{\columnwidth}{@{\extracolsep{\fill}}l c c c@{}}
\toprule[1.2pt]
Map backend & hku & ust & self \\
\midrule
MultiMap & \textbf{0.04 / 12.50} & \textbf{0.64 / 14.63} & \textbf{0.08 / 18.10} \\
ikd-Tree High & 0.44 / 16.22 & 2.21 / 19.17 & 0.64 / 22.17 \\
ikd-Tree Low & 0.70 / 12.56 & 2.12 / 16.46 & 2.75 / 20.56 \\
\bottomrule[1.2pt]
\end{tabular*}
\end{table}

\begingroup
\widowpenalty=10000
Although m3o1--m3o2 are RMSE-evaluated, their higher ratios reflect complex outdoor scenes, where DALIO attains the best or tied-best RMSE with low runtime. In Table~\ref{tab:time_eval_e2e}, higher ratios similarly indicate complex local structure, so MultiMap preserves finer anchors; together with batch selection and the conservative weak-direction update, this leads to lower E2E errors on challenging m3i, dege, and self sequences while keeping runtime low. The accuracy of DALIO\ensuremath{{}^{*}} is generally close to DALIO, confirming that batch selection does not introduce a noticeable accuracy drop while reducing redundant point processing.
\par
\endgroup

\begin{figure*}[ht]
		\centering
		\includegraphics[width=0.99\linewidth]{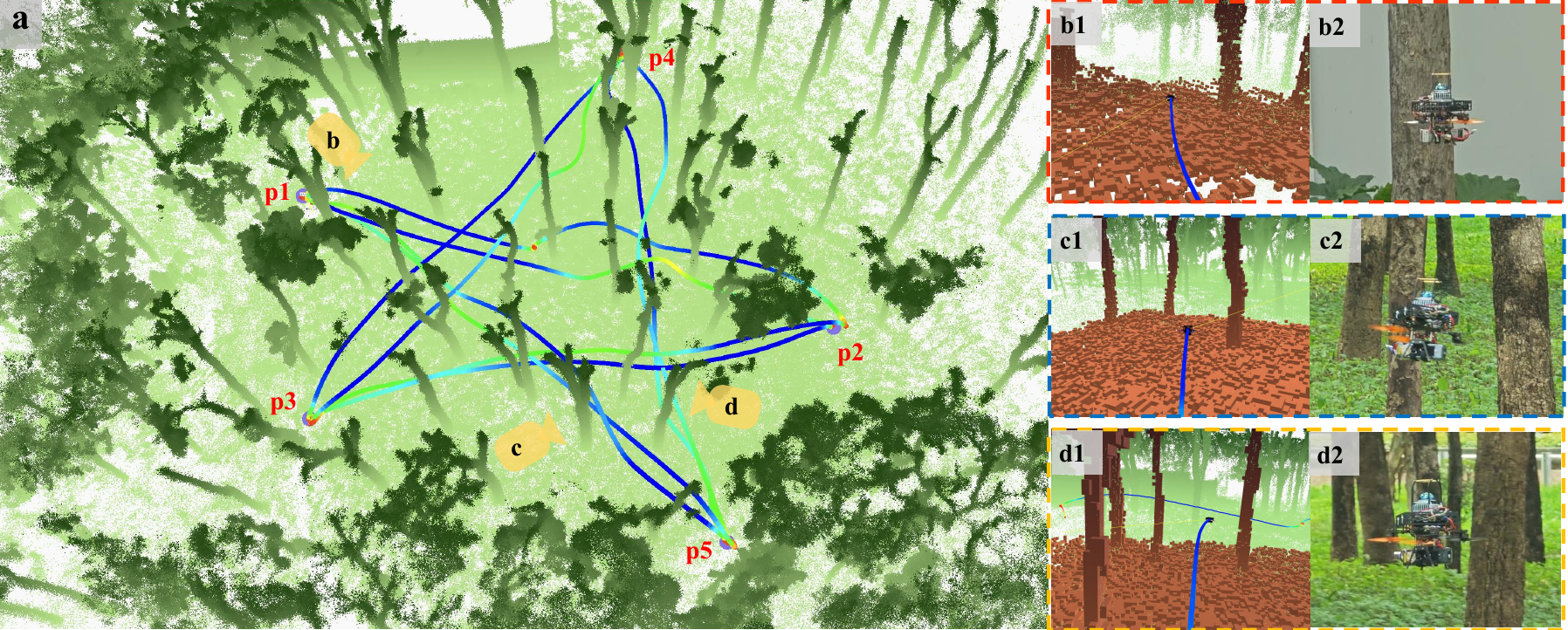}
		\caption{Fig. a shows the complete point cloud map, with the trajectory color-coded by velocity. b, c, and d present multiple flight perspectives. b1, c1, and d1 show the occupancy grid maps used for real-time obstacle avoidance, while b2, c2, and d2 display the real-world UAV flight trajectories captured by a camera.} 
		\label{fig:rl_fly} 
\end{figure*} 
\begin{figure}[h]
		\centering 
		\includegraphics[width=\linewidth]{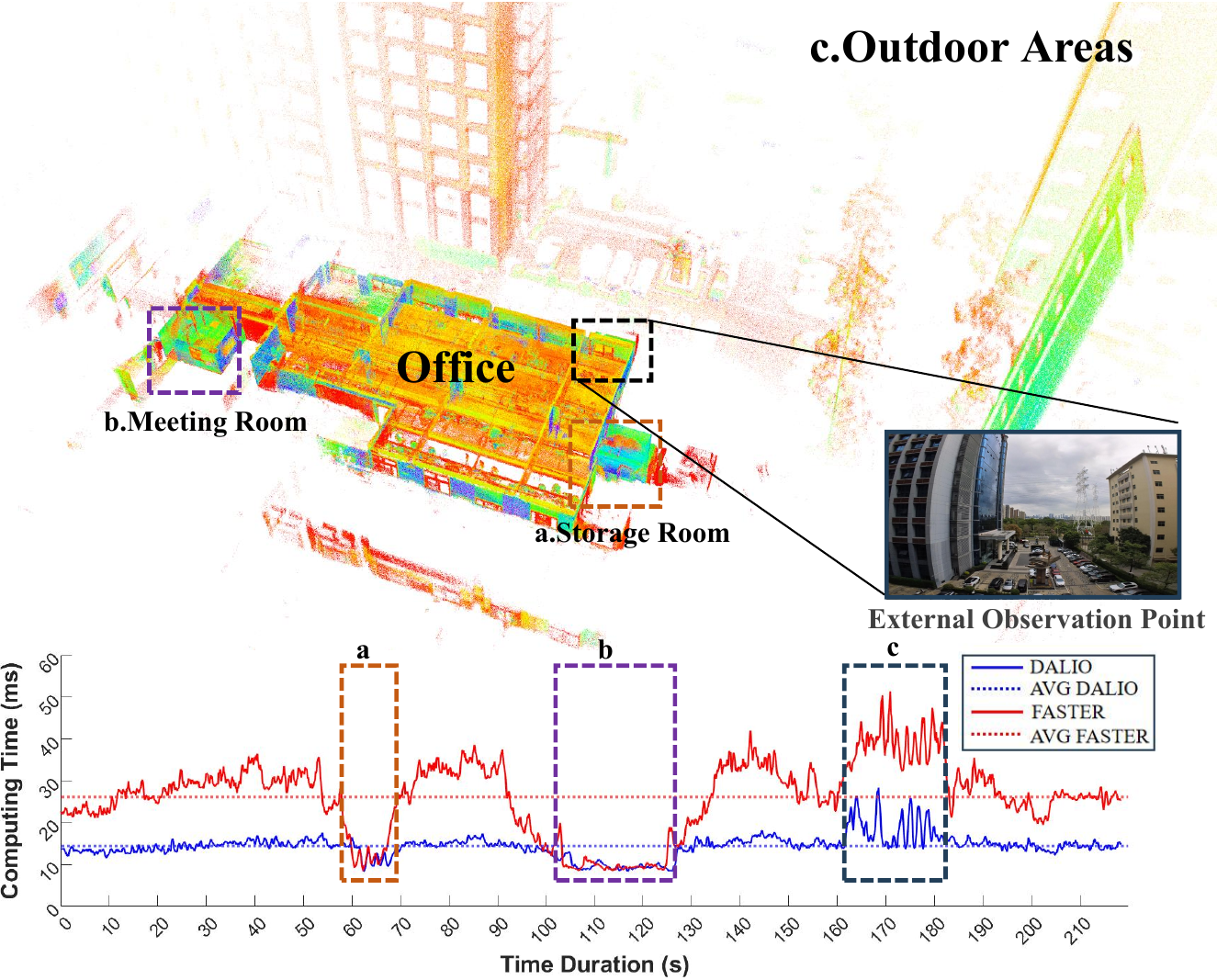}
		\caption{Running results of the algorithm on the handheld device, including mapping outputs (top) and runtime over time (bottom). Regions (a), (b), and (c) indicate different segments of the runtime curve corresponding to the device operating in various environments.}
		\label{fig:handle}
\end{figure}

\subsubsection{\texorpdfstring{Map Ablation}{Map Ablation}}
We isolate the effect of the map backend by disabling the degeneracy-aware batch selector for all variants and replacing only the map structure.
The compared backends are MultiMap with adaptive resolutions of 0.5 m--0.25 m--0.125 m, ikd-Tree Low with a fixed 0.5 m resolution, and ikd-Tree High with a fixed 0.125 m resolution.
We use hku, ust, and self-collected sequences because their indoor--outdoor transitions and local structural details directly stress the map-resolution trade-off. For MultiMap, the reported averages are the arithmetic means of the corresponding DALIO\ensuremath{{}^{*}} rows in Table~\ref{tab:time_eval_e2e}.

As shown in Fig.~\ref{fig:map_compare}, the fixed high-resolution map preserves finer local structures but causes substantially faster map-size growth, while the fixed low-resolution map is compact but may lose small-scale constraints. MultiMap avoids this fixed-resolution trade-off by keeping coarse voxels where the geometry is already stable and refining regions that need finer local anchors. The quantitative map-only ablation in Table~\ref{tab:hku_hkust_e2e_time} shows that this adaptive representation maintains lower loop-return error than the fixed-resolution maps while avoiding the runtime and memory growth of the high-resolution setting. The Split Voxel Ratio in Tables~\ref{tab:time_eval_rmse} and~\ref{tab:time_eval_e2e} follows the same behavior: small values correspond to regular scenes where coarse anchors are often sufficient, while larger values appear in mixed-scale, forest, or rotating-LiDAR scenes where MultiMap benefits from preserving finer local structure.
\subsubsection{\texorpdfstring{Degeneracy Analysis and Suppression Ablation}{Degeneracy Analysis and Suppression Ablation}}

We further analyze the degeneracy level using the number of weak dimensions in the pose-related observation matrix. In the corridor sequences, the degeneracy is mainly one-dimensional: the weakly constrained direction is along the corridor, where repeated wall and floor structures provide little geometric information for motion parallel to the corridor. In contrast, the  dege1--dege2 sequences contain limited-FOV LiDAR motion while facing a large wall-like planar surface. Such geometry can weaken multiple pose directions simultaneously, and the number of weak dimensions can reach three. To isolate the role of suppression, DALIO\ensuremath{{}^{\dagger}} removes only the observable-subspace suppression update while retaining the same batch selector and MultiMap. Table~\ref{tab:time_eval_rmse} shows that this variant stays close to DALIO on mostly well-constrained ground-truth sequences. In contrast, Table~\ref{tab:time_eval_e2e} shows clear E2E degradation on representative degenerate or locally weakly constrained sequences, including m3i1--m3i2, dege1--dege2, and self2--self4. These results indicate that the suppression update improves loop-return stability by avoiding unreliable LiDAR corrections along weakly constrained directions, especially when the scene degeneracy is stronger than a simple one-direction corridor case.
\subsection{Evaluation on Handheld Devices}

We evaluate DALIO on two handheld platforms: (1) the RK3588 handheld device described above, and (2) a custom rotating LiDAR system designed to expand the field of view (FOV) for improved mapping.

\subsubsection{ARM-Based Handheld Device}

We validated DALIO on the same RK3588-based handheld device shown in Fig.~\ref{fig:hard}a. A Livox Mid-360 LiDAR was used for sensing, while the onboard camera was used solely for recording. The system was tested across various real-world environments, including an office (medium scale), a small meeting room and a storage room (small scale), and large outdoor areas visible through windows. As shown in Fig.~\ref{fig:handle}, DALIO maintained consistently low computational cost, with minimal runtime variation when transitioning from small to large-scale environments. The generated maps were clear and globally consistent, and the end-to-end localization error upon returning to the starting point was low.

We also evaluated FASTER-LIO on the same device using identical parameters. As shown in Fig.~\ref{fig:handle}, DALIO consistently outperformed FASTER-LIO in computational efficiency while achieving better localization accuracy. In the storage room and small meeting room (Fig.~\ref{fig:handle}(a) and (b), respectively), both methods had similar runtime due to the limited number of downsampled points, with DALIO using nearly all available points to maintain robustness. In contrast, in the office and large outdoor areas, DALIO benefited from its point cloud batching strategy, requiring only a subset of points for stable localization, thus significantly reducing runtime. On average, DALIO achieved a runtime of {\SI{14.26}{ms}} per frame, compared to {\SI{27.34}{ms}} for FASTER-LIO. 

\begin{figure}[t]
	\centering
	\includegraphics[width=\linewidth]{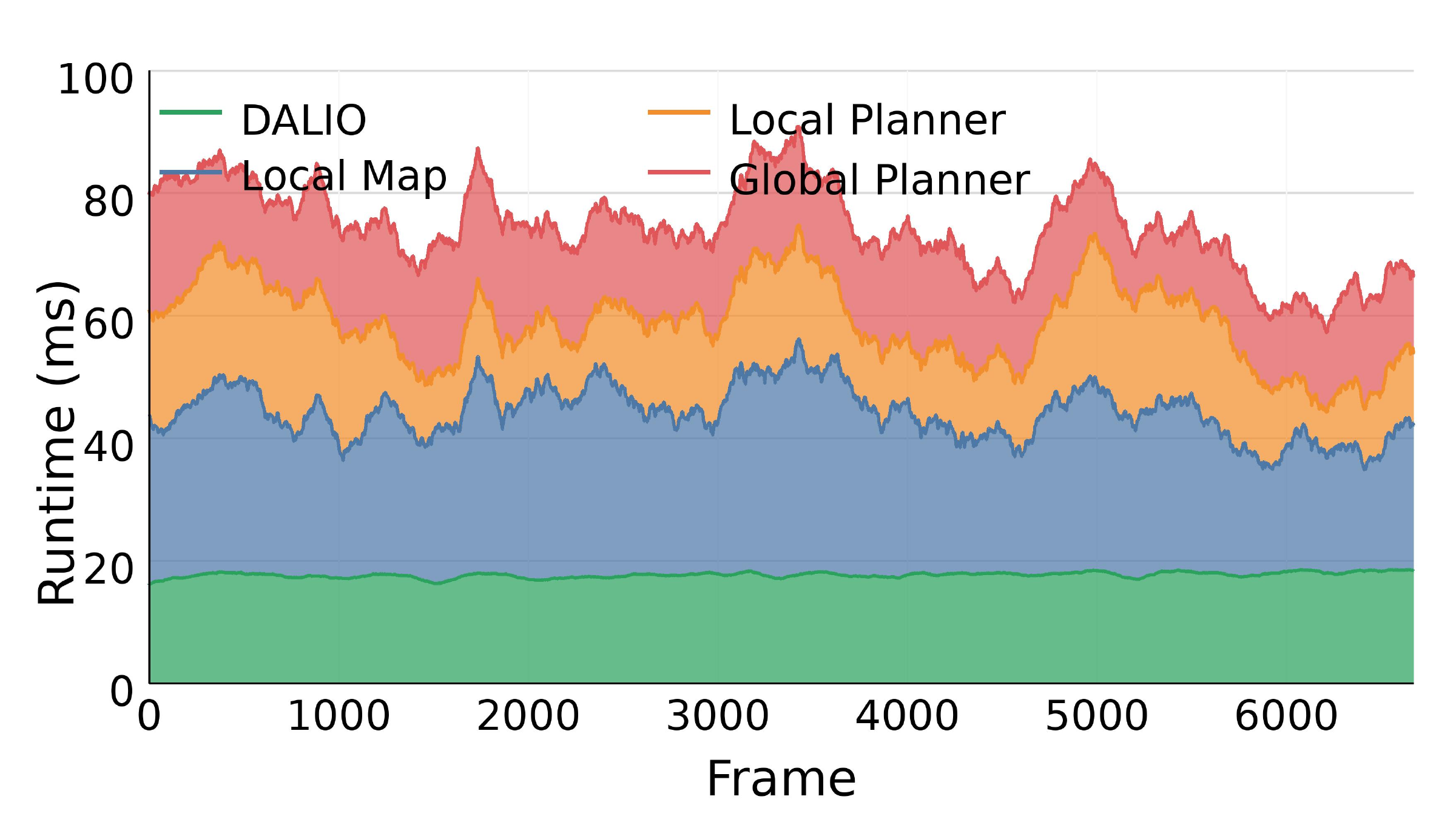}
	\caption{Runtime of DALIO, local mapping, local planning, and global planning during high-speed forest flight on the Intel i3 onboard computer.}
	\label{fig:uav_i3_full_stack_timing}
\end{figure}
\subsubsection{Rotating LiDAR Device}

To further evaluate DALIO in large-scale and challenging environments, we collected a dataset using a rotating LiDAR device that continuously spins at {\SI{1}{Hz}}. 
This setup, based on a Livox Mid-360 LiDAR and an Intel N305 computing unit, produces full-surround point clouds but also introduces additional challenges: the LiDAR orientation is uncontrollable, and scans may temporarily align with degenerate directions, especially in narrow spaces (Fig.~\ref{fig:hkuBigScene}(b1)--(b3)). 
Fig.~\ref{fig:hard}c illustrates the device configuration. 
The dataset covers a {\SI{1021}{m}} trajectory across diverse environments, including degenerate corridors, staircases, outdoor building areas, and forested scenes. 
In these scenarios, DALIO's degeneracy suppression module rejects unreliable LiDAR sensitivity along weakly constrained directions, maintaining stable localization during short-term degeneracy.

Beyond local degeneracy, this dataset also features drastic environmental variation. Fixed-resolution maps fail to provide stable localization across all conditions, whereas our MultiMap adapts its resolution online, enabling a single parameter set to handle large changes in scene scale and geometry.

The quantitative results for this rotating-LiDAR sequence are reported as self2 in Table~\ref{tab:time_eval_e2e}. DALIO achieves a low end-to-end return error with the lowest runtime on this challenging sequence, whereas several baseline methods either drift severely or fail. Compared with the all-point variant, DALIO also reduces runtime substantially while improving loop-return consistency.

\subsection{Evaluation on UAV Navigation}

In the UAV experiments, we present two application scenarios: (1) trajectory tracking in a mixed indoor--outdoor environment, and (2) high-speed maneuvering in a forest.

\subsubsection{Trajectory Tracking in Mixed Indoor--Outdoor Environments}

To verify DALIO's stability in high-speed UAV navigation through both indoor and outdoor areas, we deploy it onboard a UAV and integrate it with SUPER~\cite{ren2025super} as the online planning module.

We conduct three trajectory tracking experiments with progressively increasing flight speeds. Fig.~\ref{fig:fly} illustrates the highest-speed case, Test~3, where the UAV reaches up to {\SI{5}{m/s}}. The trajectory is color-coded by velocity, with a total flight distance of {\SI{181}{m}}. A set of predefined waypoints guides the UAV while enabling real-time obstacle avoidance. The trajectory begins indoors, transitions to a large outdoor loop, and returns indoors, demonstrating DALIO's stable and real-time localization performance across diverse flight conditions. As shown in Fig.~\ref{fig:fly}, the generated map remains consistent throughout the flight. In particular, Fig.~\ref{fig:fly}c highlights the MultiMap strategy, where high-resolution red points capture detailed indoor structures, and low-resolution green points represent large-scale outdoor mapping.

Across the three real-flight tests, the UAV returned to the starting area with consistently low end-to-end errors despite indoor--outdoor transitions and changing pedestrian or obstacle configurations.

\subsubsection{High-Speed Maneuvering in a Forest}

\begingroup
\widowpenalty=10000
To evaluate DALIO's localization performance during aggressive maneuvers, we conduct a high-speed flight in a forest. The UAV takes off within a $\SI{20}{m} \times \SI{20}{m}$ area and flies between five waypoints arranged in a pentagon-shaped pattern. The UAV departs from $p_1$, passes sequentially through $p_2$ to $p_5$, repeats the loop once, and finally lands back at $p_1$. The maximum flight speed reaches {\SI{8}{m/s}}, the peak acceleration reaches $5g$, and the maximum angular velocity reaches {\SI{10}{rad/s}}. DALIO provides accurate, real-time odometry throughout the maneuvers. To assess loop-return consistency, the UAV completes two full circuits of the trajectory, as shown in Fig.~\ref{fig:rl_fly}a. The figure also shows that UAV speed varies significantly, as it must avoid obstacles that suddenly appear in the FOV during high-speed flight without a prior map. DALIO maintains stable localization even under these highly dynamic conditions. Moreover, in such geometrically sparse environments, DALIO continues to achieve accurate localization with low computational cost, leaving sufficient resources for perception and \mbox{planning modules}.
\par
\endgroup

Fig.~\ref{fig:uav_i3_full_stack_timing} further reports the system-level runtime on the low-cost Intel i3 onboard computer used in this forest flight. DALIO runs together with local mapping, local planning, and global planning, and the stacked runtime remains within the real-time budget during the aggressive trajectory. This result shows that the reduced LIO cost is not only an offline benchmark improvement: it leaves a stable computation window for the rest of the navigation stack, allowing the UAV to maintain stable flight while performing perception, planning, and control on a resource-limited platform.

\section{Conclusion and Future Work}
This paper presented DALIO, a LiDAR--inertial odometry framework with a degeneracy-aware batch selector, a degeneracy suppression strategy, and a multi-resolution map. Experiments on diverse platforms demonstrate that DALIO achieves significantly lower runtime while maintaining competitive localization accuracy and robust performance. Future work will focus on handling dynamic environments, such as filtering moving objects during mapping, and extending the framework with semantic and global optimization modules for improved long-term consistency.

\section*{Acknowledgment}
Many thanks to our labmates Zhengzhe Xu and Kaiwen Jiang for their outstanding contributions to the Forest Flight experiments. This impressive demo wouldn't have been possible without their support.

\bibliography{paper}

\begin{IEEEbiography}[{\authorphoto{figures/authors/zhd.jpg}}]{Haoda Zhu}
received the B.E. degree in automation in 2024 from the School of Mechanical Engineering and Automation, Harbin Institute of Technology, Shenzhen, China. He is currently a Ph.D. student at The University of Hong Kong.

His research interests include robotics, sensor fusion, and LiDAR-based simultaneous localization and mapping (SLAM).
\end{IEEEbiography}

\begin{IEEEbiography}[{\authorphoto{figures/authors/zfc.jpg}}]{Fangcheng Zhu}
received the B.E. degree in automation in 2021 from the School of Mechanical Engineering and Automation, Harbin Institute of Technology, Shenzhen, China. He received the Ph.D. degree from The University of Hong Kong, Hong Kong SAR, China.

His research interests include robotics and aerial swarm systems, with a focus on LiDAR-based simultaneous localization and mapping (SLAM) and sensor calibration.
\end{IEEEbiography}

\begin{IEEEbiography}[{\authorphoto{figures/authors/fangyuhao.jpg}}]{Yuhao Fang}
received the B.E. degree from the Harbin Institute of Technology, Shenzhen, China. He is currently working toward the M.Sc. degree with the School of Intelligence Science and Engineering, Harbin Institute of Technology, Shenzhen, China.
His research interests include computer vision, robotics, and SLAM.
\end{IEEEbiography}

\begin{IEEEbiography}[{\authorphoto{figures/authors/hjl.jpg}}]{Jiale Han}
received the M.Sc. degree from Shanghai Jiao Tong University (SJTU), Shanghai, China. He is currently with the Department of Mechanical Engineering, The University of Hong Kong (HKU), Hong Kong SAR, China.

His research interests include autonomous navigation and robotic perception.
\end{IEEEbiography}

\begin{IEEEbiography}[{\authorphoto{figures/authors/ryf.png}}]{Yunfan Ren}
received the B.Eng. degree in automation from Harbin Institute of Technology, China, in 2021. He received the Ph.D. degree from The University of Hong Kong, Hong Kong SAR, China.

His research interests include autonomous navigation, UAV planning, optimal control, and swarm system autonomy. 
\end{IEEEbiography}

\begin{IEEEbiography}[{\authorphoto[1pt]{figures/authors/lsq.pdf}}]{Siqi Liang}
received the B.E. degree in automation in 2023 from the School of Mechanical Engineering and Automation, Harbin Institute of Technology, Shenzhen, China. He is currently working toward the Ph.D. degree in mechanical engineering at the University of Hong Kong (HKU), Hong Kong SAR, China.

His research interests include robotics, state estimation, and LiDAR SLAM.
\end{IEEEbiography}

\begin{IEEEbiography}[{\authorphoto{figures/authors/ylj.jpg}}]{Longji Yin}
received the B.Eng. degree in automation and the B.A. degree in English literature from Zhejiang University, Zhejiang, China, in 2019 and the M.Sc. degree in robotics from the Johns Hopkins University, Baltimore, MD, USA in 2021. He received the Ph.D. degree from The University of Hong Kong, Hong Kong SAR, China.

His research interests include motion planning and control for autonomous aerial robots.
\end{IEEEbiography}

\begin{IEEEbiography}[{\authorphoto{figures/authors/mj.jpg}}]{Jie Mei}
received the B.S. degree in information and computing science from Jilin University, Changchun, China, in 2007, and the Ph.D. degree in control science and engineering from the Harbin Institute of Technology, Harbin, China, in 2011.

He is currently a professor with the School of Intelligence Science and Engineering, Harbin Institute of Technology, Shenzhen, China.

His research interests include nonlinear control, distributed control of multi-agent systems, and their applications in robotics.
\end{IEEEbiography}

\begin{IEEEbiography}[{\authorphoto{figures/authors/fuzhang.png}}]{Fu Zhang}
received the B.E. degree in automation from the University of Science and Technology of China (USTC), Hefei, China, in 2011 and the Ph.D. degree in controls from the University of California at Berkeley, Berkeley, CA, USA, in 2015.

He joined the Department of Mechanical Engineering, University of Hong Kong (HKU), Hong Kong, as an assistant professor in 2018. He was promoted to a tenured associate professor at the University of Hong Kong (HKU) in 2024. His current research interests include robotics and controls, with a focus on UAV design, navigation, control, and LiDAR-based SLAM.
\end{IEEEbiography}

\vfill
\end{document}